\RequirePackage{fix-cm}
\documentclass[smallextended]{svjour3}       
\smartqed  
\usepackage{graphicx}
\usepackage{times}
\usepackage{epsfig}
\usepackage{graphicx}
\usepackage{amsmath}
\usepackage{amssymb}
\usepackage{array,tabularx}
\usepackage{enumitem}
\usepackage{multirow}
\usepackage{epstopdf}
\usepackage{subfigure,threeparttable}
\usepackage{lscape}
\usepackage{url}
\usepackage{soul}
\usepackage{caption}
\usepackage{lscape}
\usepackage[numbers]{natbib}
\begin{document}\sloppy

\title{A novel hybrid score level and decision level fusion scheme for cancelable multi-biometric verification}

\author{Rudresh Dwivedi         \and
        Somnath Dey 
}


\institute{Rudresh Dwivedi \and Somnath Dey \at
              Discipline of Computer Science \& Engineering, \\Indian Institute of Technology Indore\\
              Simrol, Indore, India-453552\\
              Tel.: +91-9713888726\\
              \email{phd1301201006@@iiti.ac.in}           
}

\date{Received: date / Accepted: date}

\maketitle

\begin{abstract}
In spite of the benefits of biometric-based authentication systems, there are few concerns raised because of the sensitivity of biometric data to outliers, low performance caused due to intra-class variations and privacy invasion caused by information leakage. To address these issues, we propose a hybrid fusion framework where only the protected modalities are combined to fulfill the requirement of secrecy and performance improvement. This paper presents a method to integrate cancelable modalities utilizing mean-closure weighting (MCW) score level and Dempster-Shafer (DS) theory based decision level fusion for iris and fingerprint to mitigate the limitations in the individual score or decision fusion mechanisms. The proposed hybrid fusion scheme incorporates the similarity scores from different matchers corresponding to each protected modality. The individual scores obtained from different matchers for each modality are combined using MCW score fusion method. The MCW technique achieves the optimal weight for each matcher involved in the score computation. Further, DS theory is applied to the induced scores to output the final decision. The rigorous experimental evaluations on three virtual databases indicate that the proposed hybrid fusion framework outperforms over the component level or individual fusion methods (score level and decision level fusion). As a result, we achieve (48\%,66\%), (72\%,86\%) and (49\%,38\%) of performance improvement over unimodal cancelable iris and unimodal cancelable fingerprint verification systems for Virtual\_A, Virtual\_B and Virtual\_C databases, respectively. Also, the proposed method is robust enough to the variability of scores and outliers satisfying the requirement of secure authentication.
\end{abstract}

\section{Introduction}
Over the last decade, reliable and accurate biometric-based authentication is the utmost need for numerous applications such as access control, banking, and healthcare \cite{pedia}. Despite several benefits, there are various privacy and performance issues exist such as noisy data, sensitivity to outliers, lack of uniqueness, identity invasion and absence of revocability in unimodal biometric systems \cite{handbook}. In this regard, security and performance are the two key parameters for recognition systems. As a result, there is a need for a biometric system which can achieve template protection as well as performance improvement to yield robustness to the system \cite{mulp}. For protecting original biometric template, the idea of cancelable biometric is introduced with three requirements: non-invertibility, diversity and revocability \cite{btpnew}. Non-invertibility states that the original biometric template should not be recovered from the protected one. Diversity refers to derive numerous templates from the same original template whereas revocability denotes that a new template must be issued if a stored protected template gets compromised.  These three criteria must be satisfied to ensure the privacy of any user and to prevent any possibility of security theft \cite{mulp,fupro}. To improve the performance, biometric fusion \cite{handbook} is adopted in recent years which combines the information from more than one biometric modality. Therefore, we focus on multibiometric fusion to ascertain improvement in the performance and utilize cancelable biometric to assure privacy protection to alleviate the aforementioned limitations.

There exist five different level of fusion i.e. sensor, feature, rank, score, and decision. Most of the feature level multimodal protection approaches \cite{paulm,mulp,mixross,towards,printvein,chin} involve concatenation, random projection or transformation based on a user-specific PIN for privacy protection. The approaches proposed in \cite{paulm,mulp,mixross,towards,printvein,chin} lead to a minor performance improvement over the unimodal biometric system. Moreover, if a protected multi-biometric template gets compromised, there is no possibility to prevent the loss of original biometric information. The existing works in \cite{camli,sutcu} choose fingerprint as a biometric modality and incorporate feature-level fusion. Hence, there is no mode present to re-transform it with a new key if multi-biometric template gets compromised.  Though the methods proposed in \cite{cimato,deepnn} shows a significant performance improvement, they are too complex for real-world implications. Besides feature level, there have been various methods \cite{lmezai,kabir,acoajay,scorelevel,ds1} incorporating score fusion in recent years. Mezai et al. \cite{lmezai} applied score fusion over face and voice biometric using DS theory. The method proposed by Mezai et al. \cite{lmezai} performs poor in verification scenario. Recently, Kabir et al. \cite{kabir} proposed confidence-based weighting (CBW) for score fusion after normalization. This weighting scheme does not utilize the complete set of scores which aids to uncertainty. Kumar et al. \cite{acoajay} applied Ant Colony Optimization (ACO) to evaluate weights in sum-rule based score fusion. The limitation lies in the complex implementation and it may also suffer from the local minimum problem. Sadhya et al. \cite{softbio} utilized Bayesian decision theory to combine soft biometric traits. In this method \cite{softbio}, the performance degrades if matcher's characteristics or distributions are not known. Nguyen et al. \cite{ds1} proposed DS theory based score fusion for iris and fingerprint. The method provides an incorrect decision in scenarios where sufficient training samples are absent.

Among the five different fusion levels, score level fusion is favored owing to the factors such as ease of fusion and freedom to choose any feature extraction and matching algorithms \cite{joshi}. Despite many benefits, many commercial firms provide access only on the basis of the final decision or recognition output. Further, if the involved matchers are non-homogeneous or do not have the same scale, score level fusion becomes a challenging task. Score fusion normalizes individual scores from classifiers/ matchers and returns combined score whereas decision fusion returns a final decision by combining individual decisions from classifiers/ matchers. Hence, we have chosen score level as well as decision level fusion which would overcome the limitations of the score as well as decision fusion if a combination of both fusion mechanism is employed. The notable motives behind this work are to integrate multiple biometric modalities thereby achieving more discrimination among different users and producing significant improvement in (i) overall accuracy/ performance, (ii) Flexibility owing to the absence of individual training (iii) Avoiding spoof attacks. As a multimodal biometric system is able to attain performance improvement and overcome spoofing attacks, template security schemes aim to protect original biometric data. The compositions of these two schemes achieve a win-win scenario for the template protection and performance enhancement. Out of the two mechanisms of template protection named cancelable biometric and biometric cryptosystems, we utilize cancelable transformation instead of biometric cryptosystem since cryptographic methods reform the template and generate a poor match score. Owing to the above discussion, we proceed in the direction of hybrid fusion integrating both score and decision level mechanisms to overcome the limitations related to unibiometric, multibiometric, protected multibiometric systems as described above.

In our prior work \cite{scorelevel}, we have introduced a two-level score fusion scheme which utilizes the difference between the mean of distributions and scores values to evaluate the weight to combine different matchers corresponding to each modality for the first level. In the next level, the rectangular area is considered for integrating different modalities. However, the limitations of our earlier work lie in the aspect of performance and security. Moreover, the prior work does not perform well if non-homogeneous matchers are present for different modalities in any biometric authentication system. In a nutshell, we extend this work in the following aspect as compared to our previous work. In this work, we utilize our previous works \cite{myiris,myfinger} to derive cancelable template for iris and fingerprint. Next, we apply the mean-closure weighting (MCW) to combine scores from individual matchers corresponding to one particular biometric modality. Finally, DS theory of evidence is employed to combine the fused scores achieved from different biometric modalities. For applied intelligence perspective, DS theory based fusion is utilized to mitigate the uncertainty associated with non-homogeneous matcher's output. Further, the proposed hybrid fusion technique can be efficiently applied for verification decision making in security infrastructure, defense, governments, and industries. To the best of our knowledge, our method is the first to incorporate the hybrid fusion for protected multimodal verification utilizing MCW based score level and DS theory-based decision fusion.

Despite many efforts in the past, existing works suffer from various limitations. The main contributions of this work are highlighted with respect to the limitations present in the state-of-the-art:
\begin{enumerate}
	\item We have proposed a generic hybrid fusion framework for combining multiple protected biometric modalities (iris and fingerprint) to aid security and improve performance.
	
	\item We have chosen score level fusion at the first level to combine scores from homogeneous matcher's scores as well as decision level fusion at the second level to integrate different modalities which would overcome the limitations of individual score fusion as well as decision fusion if a combination of both fusion mechanism is employed.
	
	\item In this work, score fusion is applied directly onto raw scores which contain richer information than the normalized scores. No normalization is required as the scores evaluated from different matchers are already obtained in the interval of [0,1].
	
	\item We perform two-level fusion onto cancelable (protected) scores utilizing mean-closure and DS theory method to reduce the performance degradation due to cancelable transformation.
	
	\item Experimental evaluation onto three different virtual databases is carried out to explore the potential robustness of the proposed method for multimodal biometrics. Also, we have compared our method with state-of-the-art fusion methodologies. We have also compared our method with the recent fusion methods including individual feature, score and decision fusion methods. The experimental results conclude that our method outperforms over the existing feature, score, decision and hybrid fusion approaches in the literature.
	
	\item We have computed the performance in two scenarios; (i) performance with fused scores obtained from cancelable templates (ii) performance with fused scores obtained from unprotected (original) templates. The experimental evaluation confirms a minor performance degradation with respect to the unprotected multibiometric system. Also, we achieve a significant relative performance improvement over the unimodal cancelable biometric systems.
	
	\item We have conducted security analysis concerning the required criteria of non-invertibility, diversity, and revocability using hybrid fusion security model.
	
	\item The proposed hybrid framework has mitigated the uncertainty present in the biometric data and proved to be less complex in comparison to other hybrid fusion methods.
\end{enumerate}
The rest of this paper has the following structure. In Section 2, existing approaches on cancelable multibiometric are discussed. Section 3 briefly describes DS-theory of evidence. In Section 4, the proposed method is discussed. Experimental evaluations are presented in Section 5, and Section 6 discusses the security analysis. The conclusions are drawn in Section 7.

\section{Related works}
Utilization of just one biometric modality may result in performance degradation and security invasion. Hence, multi-biometric authentication systems allowing template protection have become a promising solution to address these concerns. Also, the utilization of multimodal biometric system would aid to compensate the performance drop caused due to cancelable transformation. Therefore, we discuss the existing works in context to cancelable multimodal biometric systems in this section. 
In literature, various methods \cite{paulm,mulp,mixross,camli} have been proposed to provide secure authentication for multi-biometric systems. Paul et al. \cite{paulm} evaluated the unique features from face and ear. After dimension reduction through PCA, they applied random projection to protect the multi-biometric template. Canuto et al. \cite{mulp} utilized different decision fusion methods by applying biohashing, interpolation, and bioconvolving based protection mechanism over voice and iris biometric. Othman et al. \cite{mixross} acquired two fingerprint impressions from two different fingers of a user and combined spiral and the continuous components to create a new protected identity. Camlikaya et al. \cite{camli} incorporated feature level fusion over voice and fingerprint by embedding the minutiae information into the computed voice features for privacy preservation. In recent years, Rathgeb et al. \cite{towards} proposed Bloom filter based integration scheme to combine face and iris features. Yang et al. \cite{printvein} proposed a method to integrate fingerprint and finger-vein features thereby improving recognition accuracy and security. They evaluated invariant features and concatenated the features after dividing the features into blocks. Finally, Partial-DFT is performed on concatenated features to derive multibiometric cancelable template. Chin et al. \cite{chin} propose a feature-level fusion scheme onto fingerprint and palmprint where Gabor filtering is applied to evaluate features. Next, random tiling is used to derive multimodal protected template based on a user-specific PIN. Kelkboom et al. \cite{busch} applied feature, score and decision level fusion strategies over helper data. Helper data consists of reliable bits derived from face database. Next, they utilized AND rule and OR rule for all individual fusion. 

Besides cancelable transformation, many researchers \cite{sutcu,deepnn,cimato} also applied fusion schemes over biometric cryptosystem based transformations.  Sutcu et al. \cite{sutcu} evaluated feature string for face and fingerprint biometric and fed it as the input to a fuzzy commitment framework after concatenating both strings. Talreja et al. \cite{deepnn} introduced a secure multibiometric framework which evaluates features from face and iris using a Deep neural network, binarize the features and extract reliable bits. Next, error-correction is applied to the combined reliable bits, and the hash of the error-corrected bits is stored as a cancelable template. Cimato et al. \cite{cimato} introduced an approach for biometric cryptosystem where secure sketches and fuzzy extractors are applied over iris and fingerprint. Next, the hash of one modality is utilized to secure the second modality. 

Besides cancelable and cryptosystem based multi-biometric techniques, few recent works are also reported in the literature. Mezai et al. \cite{lmezai} performed score level fusion utilizing Denoeux model to transform the match scores into belief assignments. Finally, DS theory and proportional conflict redistribution (PCR) rules of combinations are applied to integrate the belief assignments. Nguyen et al. \cite{ds1} proposed a score fusion approach where uncertainty mass is computed by a linear combination of the quality scores and EER of the classifier. Further, min-max normalization is performed followed by DS theory fusion. Kabir et al. \cite{kabir} proposed two normalization methods which are overlap extrema based min-max and mean-to-overlap extrema schemes to normalize the scores. Next, a novel confidence based weighting is utilized for score integration. Kumar et al. \cite{acoajay} investigated ACO to evaluate weights for different biometric modalities taking part in the fusion. The four fusion rules are considered for combination i.e. sum, product, exp and tanh. Further, the appropriate weights for these four rules are evaluated for score fusion. Sadhya et al.  \cite{softbio} considered four soft biometric traits for combination i.e. height, weight, age, gender. Further, they applied Bayesian classifier with modified conditional probability function. Gaussian probability function and log based weighted fusion are used to achieve better performance. Tao et al. \cite{hybrid2} introduced a unique way of hybrid fusion by integrating multiple ROC (Receiver Operating Characteristics) curves. Next, the overall detection rate for all classifiers is evaluated using AND rule and OR rule. Thereafter, Grover et al. \cite{grover} proposed a hybrid fusion approach where the error rates are integrated for different threshold points using the PSO algorithm. The error rates for different threshold values are converted into the fuzzy sets using triangular membership functions. Further, global fuzzy error rates are computed by utilizing total distance criterion (TDC).

\section{Preliminaries on Dempster-Shafer theory of evidence}  
Consider, $\theta$ be a finite set of all possible hypotheses known as a frame of discernment. The power set $2^{\theta}$ contains all subsets of $\theta$ including a null set ($\phi$) and itself. Each subset in the power set is referred as a focal element and assigned a value in between [0, 1] on the basis of their evidence. A value of 1 corresponds to total belief and 0 for no belief. In general, the assigned value is named as basic belief assignment (BBA). In DS theory \cite{dec}, BBA is assigned to each subset i.e. hypothesis also called as the mass of the individual proposition,
\begin{equation}
m : 2^{\theta}\rightarrow \left [ 0,1 \right ].
\end{equation}

If $\theta =\left \{ A,B \right \} \ \text{then} \ 2^{\theta}=\left \{\varnothing ,A,B,\theta \right \}$.
The mass function fulfills the following criteria:
\begin{equation}
\sum_{a_{i}\in 2^{\theta}}m\left ( A_{i} \right )=1   , \ \ \ m\left ( \varnothing  \right )=0
\end{equation}

where $\varnothing$ represents the empty set. The measure of belief is defined by the function $bel:2^{\theta}\rightarrow \left [ 0,1 \right ]$, 
\begin{equation}
bel\left ( A \right )=\sum_{B\subseteq A, B\neq \varnothing } m\left ( B \right ).
\end{equation}
The $bel$ can also be formally defined as:
\begin{equation}
bel_{Y,t}^{\theta,\Re}\left [ E_{Y,t} \right ]\left ( w_{0}\in A \right )=x
\end{equation}
This means the degree of belief $x$ for the classifier $Y$ at time $t$ when $ w_{0}\in A $. Here, $E_{Y,t}$ represents the evidential information known to classifier $Y$ at time $t$. For ease in representation, we use $bel(A)$ instead of $ bel_{Y,t}^{\theta,\Re}\left [ E_{Y,t} \right ]\left ( w_{0}\in A \right )$. Next, plausibility ($pl$) is measured as:
\begin{equation}
pl:2^{\theta} \rightarrow \left [ 0,1 \right ], \  \  \  \  pl\left ( A \right )=1-bel(\neg A)=\sum_{B\cap A \neq \varnothing } m\left ( B \right )
\end{equation}

If $\theta$ defines the set of all possible hypotheses, then the level of uncertainty is denoted by $m\left (\theta \right)$. In a hypothesis, beliefs and disbeliefs may not sum to 1 and may attain 0 value. A value of 0 signifies no evidence present for the hypothesis.  The DS theory based aggregation involves the following steps:
\begin{itemize}
	\item The measure of belief is evaluated based on the facts from the different sources of information. As compared to Bayesian theory, the masses are not distributed among classes.
	\item Dempster rule of combination is applied to aggregate belief measure obtained from the available information and facts.
\end{itemize}

For different sources, $\left ( 1,2, \cdots, N \right )$, Dempster's rule of combination is described in Eq. (6):
\begin{equation}
m_{1,2, \cdots, N}\left ( A \right )=\frac{\sum_{B_{i}\cap \cdots \cap B_{k}=A} m_{1}\left ( B_{i} \right ) \cdot \dots \cdot m_{N}\left ( B_{k} \right )}{1-K}
\end{equation}
where $A, B_{1},\dots, B_{N}\subseteq\theta, \  \text{and}$
\begin{equation}
K= \sum_{B_{i} \cap \dots \cap B_{k}=\varnothing} m_{1}\left ( B_{i} \right ) \cdot  m_{2} \left ( B_{j} \right ) \dots m_{N} \left ( B_{k} \right )
\end{equation}
where $K$ denotes the conflict present between evidences; 1-$K$ is the normalization factor.

\subsection{Updation of masses}
In a majority of the scenarios, mass updation is required if any new evidence or belief is encountered. Suppose, $ E\subset \theta$ and $E_{d}$ be the evidence not present in $E$. If this new evidence provides the exact value of  $E_{d}$, then bel(A) is updated based on the following condition rule:
\begin{equation}
bel[E_{d}](A)=bel(A \cup \neg E)- bel(\neg E)
\end{equation}
After the computation of the masses, the classification is performed onto the training set. One of the aggregation rules is applied to evaluate total conflicting mass. Next, the winner-take-all assignment is utilized to compute $A(k)$, which is defined in Eq. (9):
\begin{equation}
m\left ( A_{k} \right )= \text{max}_{A_{j}} m\left ( A_{j} \right ), \ \ \ j=1,\dots M+1
\end{equation}
where $M+1$ represents is the total number of classes including the class of rejection and $A_{M+1}=\theta$. 

\section{Proposed method}
In this section, we describe our method to derive a hybrid fusion mechanism for cancelable multimodal biometric verification. The block diagram of the proposed fusion framework is illustrated in Fig. 1.
\begin{figure*}[!htbp]
	\centering
	\includegraphics[width=\textwidth]{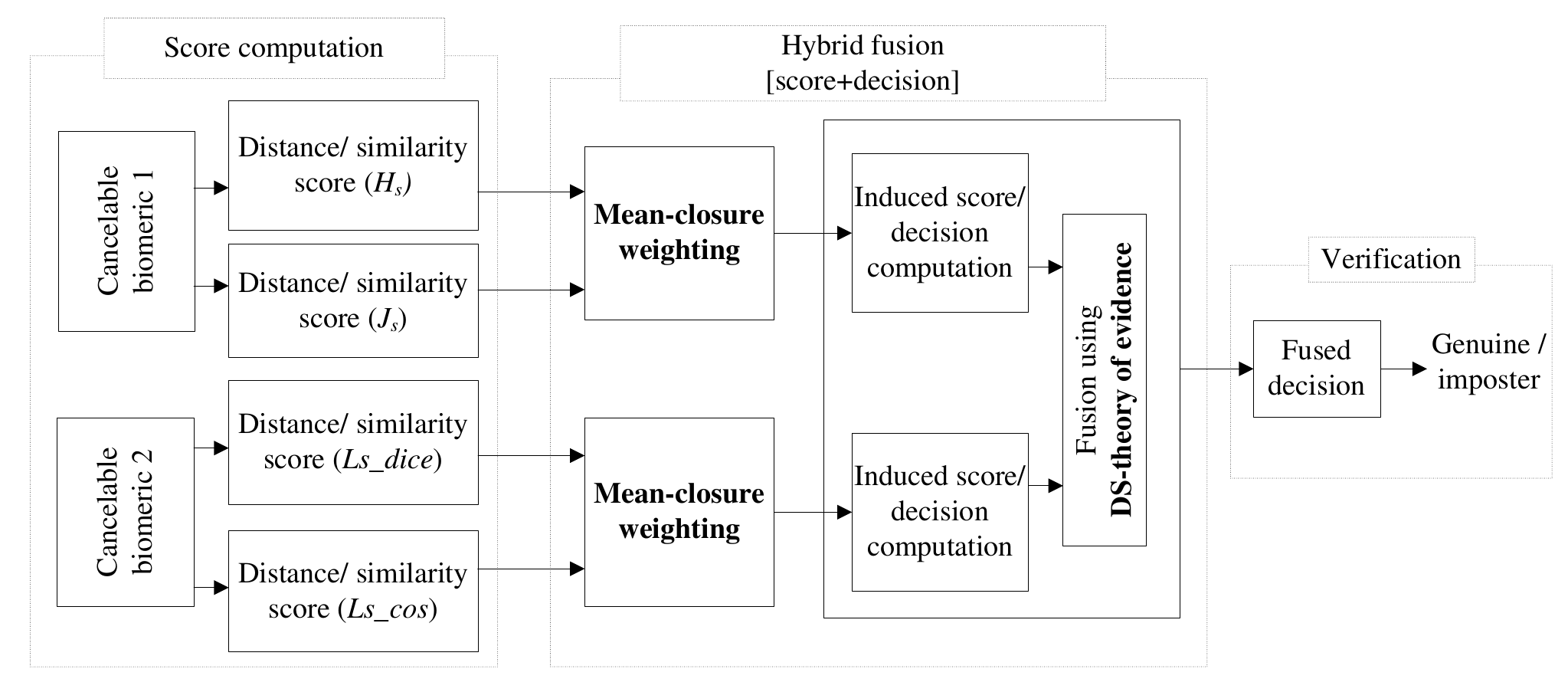}
	\caption{Block diagram of the proposed fusion framework}
	\label{fig1}\end{figure*}

The block diagram comprises three major blocks i.e. score computation, hybrid fusion, and verification. Score computation evaluates score from the different matchers applied on the protected iris and protected fingerprint templates. Hybrid fusion module includes score level followed by decision level fusion schemes. Score fusion is carried out using MCW to combine different matchers corresponding to each biometric modality whereas decision fusion integrates the decision outcome of individual modality. A final decision is made using the fusion rules for authenticating a user. Finally, verification is performed using a pre-defined threshold to classify user either a genuine or an imposter.
\subsection{Score computation}
In this subsection, we present different matchers applied on the protected templates to compute match scores. Score computation represents either dissimilarity (distance) or similarity measure. Therefore, it is needed to convert all the scores alike. In this work, we transform all the scores into similarity measure following the common standard. The comparison between cancelable enrolled and cancelable query template is performed to evaluate match scores for iris and fingerprint biometric. To derive cancelable templates for iris and fingerprint, we apply the similar methodology proposed in our earlier work \cite{myiris, myfinger}. For reader's clarity, we briefly describe the template generation methods for both modalities.

\subsubsection{Cancelable iris score computation}
To derive cancelable iris template, iris images are pre-processed using Masek's \cite{identify} and Daugman's \cite{how} techniques. Next, IrisCode features are extracted in the form of a 0-1 matrix using 1-D Log-Gabor filter \cite{identify} with phase quantization from the pre-processed iris images. Thereafter, rotation-invariant IrisCode is generated from the original IrisCodes, and the rotation invariant IrisCode is transformed into a row vector. Next, the decimal vector is derived by partitioned the row vector into fixed size blocks. Finally, a Look-up table is created to map the decimal-encoded vector and check bits are selected to generate the cancelable template. Comparison between protected enrolled and protected query iris template is performed in the transformed domain to measure the match score. First, we compute the similarity in Hamming domain for its simplicity in the evaluation. Hamming distance is the sum of non-equivalent bits (exclusive-OR) between stored and query templates. The Hamming similarity ($H_{s}$) is computed by subtracting normalized Hamming distance by one, as defined in Eq. (10):
\begin{center}
	\begin{equation}
	H_{s} = 1-\frac{1}{N}\sum_{i}^{N} E_{i}\oplus Q_{i}
	\end{equation}
\end{center}
where, $Q_{i}$ and $E_{i}$ are the $i^{th}$ bits of the query and enrolled templates, respectively. $N$ is the total number of bits in the template.

Next, Jaccard similarity is evaluated between the protected enrolled and protected query iris template. It measures the overlapping bits in $E$ and $Q$ except 0-0 overlap as defined in Eq. (11). It is computed to elude the ill match condition (0-0 match) between the protected query and protected enrolled templates.
\begin{center}
	\begin{equation}
	J_{s} = \frac{N_{11}}{N_{01}+N_{10}+N_{11}}	
	\end{equation}
\end{center}
where, \\
$N_{11}$: Number of positions where $E$ and $Q$ both have a value of 1, \\
$N_{01}$: Number of positions where the value in $E$ is 0 and the Value in $Q$ is 1,   \\
$N_{10}$: Number of positions where the value in $E$ is 1 and the Value in $Q$ is 0.

\subsubsection{Cancelable fingerprint score computation}
To derive cancelable fingerprint template, the input fingerprint image is preprocessed to obtain the thinned image and to extract the minutiae information. Next, we form a nearest-neighbor structure around each minutiae point using the ridge-based co-ordinate system and compute the ridge features from the thinned image and minutiae information. Thereafter, we apply cantor pairing function to encode the ridge features uniquely. Finally, the random projection is applied to the paired output to derive the protected template. In the verification stage, the same procedure is followed to generate the protected template from the query fingerprint. Cancelable enrolled, and cancelable query fingerprint templates are compared to calculate match scores for fingerprint biometric. The matching is performed in the transformed domain to maintain secrecy. We adopt the inner product based similarity measures since similarity computation requires measuring the likelihood between the rows in the protected enrolled template ($E$) to the rows in protected query templates ($Q$). First, we utilize Dice coefficient to measure the local similarity ($Ls\_dice$) between each row of enrolled and that of query templates as utilized in \cite{mlce} as defined in Eq. (12):
\begin{equation}
Ls\_dice \left. ( i,j \right. ) = \frac{2 E(i,:) \cdot Q(j,:)}{\left. || E(i,:) \right. ||^{2}+ \left. || Q(j,:) \right. ||^2} 
\end{equation}

Further, we apply cosine similarity ($Ls\_cos$) between each row of the enrolled and that of query templates to compute normalized dot product as defined here:
\begin{equation}
Ls\_cos \left. ( i,j \right. ) = \frac{E(i,:)\cdot Q(j,:)}{\sqrt{\left. || E(i,:) \right. ||^{2}} \sqrt{\left. || Q(j,:) \right. ||^{2}}}
\end{equation}

Next, we re-evaluate each element in local similarity matrix to avoid double matching. For this purpose, we acquire those positions where the maximum scores in $E(i,:)$, and $Q(j,:)$ coincides to obtain filtered similarity matrix. Next, the global similarity score is obtained by summing up the entries in filtered matrix and dividing by a minimum of the minutiae points in $E$ and $Q$. Finally, the likelihood of the enrolled and query template being the two fingerprint of the same subject is measured to compute global similarity scores.
\subsection{Hybrid score and decision level fusion}
After, evaluation of match scores from cancelable modalities, there is a need for score normalization such that match scores are transformed into a common interval (e.g., in the interval of [0,1]). In this work, normalization is not required since the methods utilized in score computation generate the scores in the interval of [0,1]. However, the proposed work can be extended to the situations where the scores from different biometric modalities follow different distribution range or scores derived from different matchers may have a different range instead of [0,1]. In these situations, we utilize the RHE normalization \cite{rhe} to guarantee a meaningful score integration since it is found to be sensitive to outliers. RHE minimizes the score-sets to be normalized since raw scores have richer information content than the normalized score. In the following, we have utilized the mean-closure (MC) weighting scheme for optimal weight estimation. The proposed model achieves the optimal weights for different matchers corresponding to each of the regions present in the FMR/FNMR curve including the uncertainty region. Further, the fused match scores is then fed to DS theory based decision fusion to integrate different modalities. In this work, we evaluate the scores from protected iris and protected fingerprint biometric to explore the potential significance of cancelable multimodal biometrics concerning security, privacy and performance improvement over the unimodal biometric system. 

\begin{figure}[!htp]
	\centering
	\includegraphics[height=7cm,width=\textwidth]{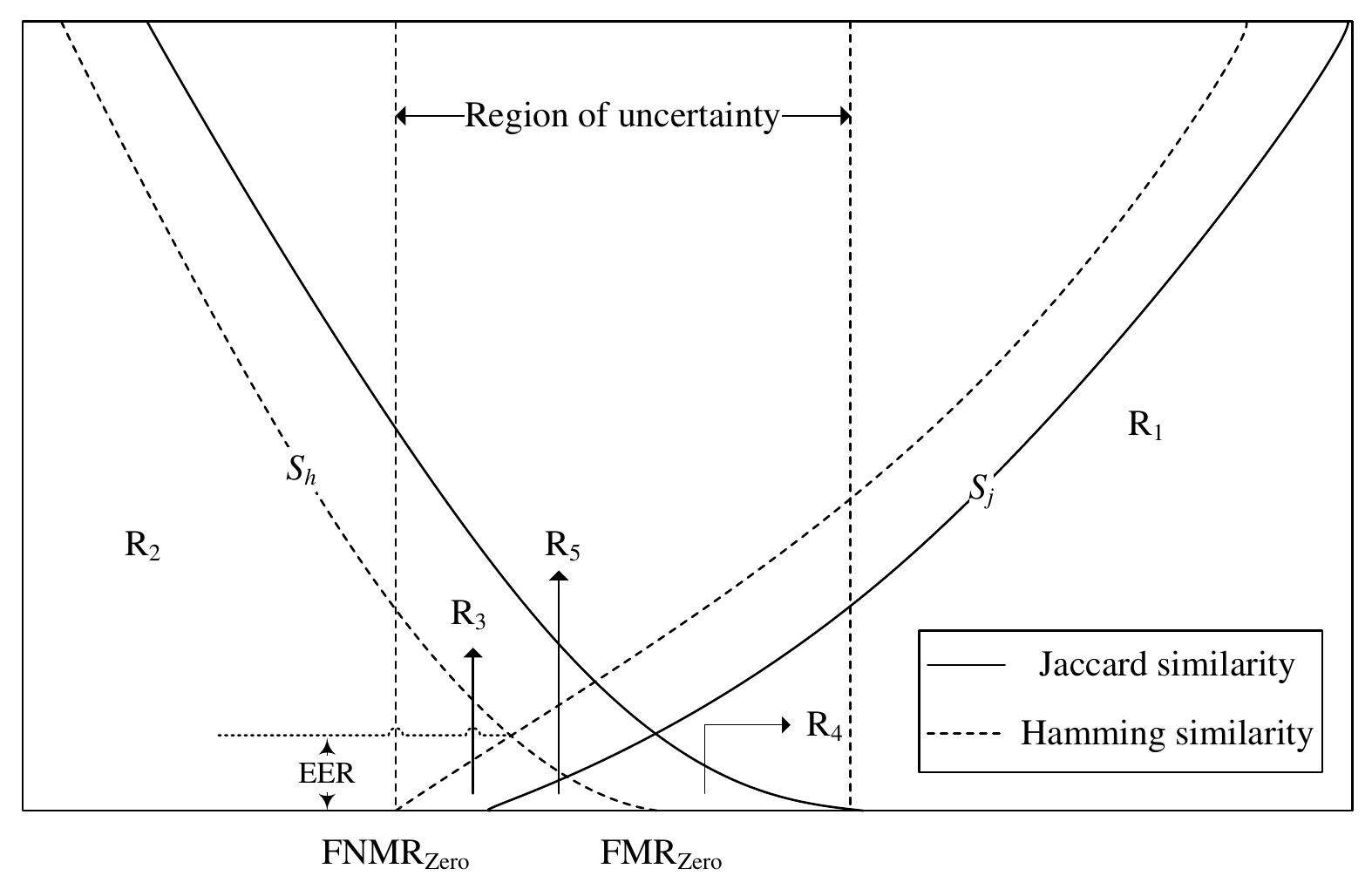}
	\caption{Explanatory diagram Region of uncertainty present in FMR/FNMR curve; $FMR_{Zero}$, $FNMR_{Zero}$ and $EER$ correspond to matcher 1 i.e. Hamming similarity}
	\label{fig2}\end{figure}

\subsubsection{Mean-closure (MC) based weighting}
Let us consider, Hamming similarity and Jaccard similarity measure to be matcher 1 and matcher 2, respectively where the scores from two matchers ($s_{h}$, $s_{j}$) are to be integrated. On the basis of Fig. 2, we indicate five possible regions for the different scores of any matcher. The regions (i.e., $R_{1}$ and $R_{2}$) represents the region of confidence where both the regions are able to classify the scores accurately. Region 3 ($R_{3}$), region 4 ($R_{4}$) and region 5 ($R_{5}$) falls into the uncertainty where it is very difficult to classify the match score.
Therefore, it is necessary to assign more weight to the scores lying into the confidence region (i.e., $R_{1}$ and $R_{2}$) and relatively less weight to the scores in the region of uncertainty while evaluating the fused score. In this work, we estimate the weights based on the mean-closure metric which measures the separation of scores from the mean of the matcher's genuine and imposter distribution for different users. The ratio of these two decides whether the user's score is close to genuine or imposter distribution of matcher 1 or matcher 2. We represent these notations as $\left ( i,m \right )$ for every pair of user and matcher. The mean-closure ($Mc_{i}^{m}$) for a of user-matcher pair $\left ( i,m \right )$ in a multibiometric system is defined as in Eq. (14):

\begin{equation}
Mc_{i}^{m}= \left (\frac{\mu _{i}^{m} \left ( gen \right ) - s_{m}}{\mu _{i}^{m} \left ( imp \right )-s_{m}}  \right )^{2}
\end{equation}
where, $\mu _{i}^{m} \left ( gen \right )$ and $\mu _{i}^{m} \left ( imp \right )$ represents the mean of genuine distribution and mean of imposter distribution, respectively. Further, the estimated weight for each matcher using MC weighting is computed as follows:

\begin{equation}
w_{i}^{m} = \frac{mc_{i}^{m}}{\sum_{i=1}^{M}mc_{i}^{m}}
\end{equation}
where, $w_{i}^{m}$ is the weight for matcher $m$ and $M$ is the number of matchers for a particular modality. $0\leq w_{i}^{m}\leq 1$, $\forall i$, $\forall m$, \text{and} $\sum_{m=1}^{M}w_{i}^{m}=1, \forall i$.

Here, the weights are proportional to the corresponding mean-closure i.e., the more accurate matcher attains higher weights than those of less accurate matcher for user $i$. This user-specific score weighting scheme deals optimal with the scores lying in the region of uncertainty. Applying the weights, we achieve fused scores from different matchers of a particular modality.

\subsubsection{Fusion using DS-theory of evidence}
In the proposed fusion framework, DS theory \cite{shafer,smets} is applied to combine the matcher's decision of individual biometric modalities. For each input image, the matchers assign either a label accept i.e., 1 to the hypothesis $i$, $i\in \theta$ or reject i.e., 0. Hence, there are two focal elements for each matcher $i$ and $\neg i=\theta-i$, where $i$ is for confirming the hypothesis and $\neg i$ is for rejecting a particular hypothesis for mass assignment as shown in Table 1. We compute the corresponding predictive rates for every matcher, which are then used to assign their BBA. The predictive rate of a matcher $P_{k}$ for an output class $k$ is the ratio of the number of users classified correctly to the total number of users classified as class $k$.

\begin{table}[!htbp]
	\centering
	\caption{Basic belief assignment function}
	\label{my-label}
	\begin{tabular}{|c|c|c|}
		\hline
		\multirow{2}{*}{Matchers} & \multicolumn{2}{c|}{Basic belief assignments (BBA)} \\ \cline{2-3} 
		& \begin{tabular}[c]{@{}c@{}}Class: Accept\\  (Gen)\end{tabular} & \begin{tabular}[c]{@{}c@{}}Class: Reject \\ (Imp)\end{tabular} \\ \hline
		Matcher 1 & $m_{1}(Gen)$ & $m_{1}(Imp)$   \\ \hline
		Matcher 2 & $m_{2}(Gen)$ & $m_{2}(Imp)$   \\ \hline
	\end{tabular}
\end{table}

After applying the MCW method to combine score from the different matchers for a particular modality, we utilize DS theory of evidence to integrate the scores from different modality to obtain the overall score/decision. For this purpose, we evaluate decision induced scores from the fused score and apply DS theory framework to obtain a final decision output. In the proposed method, when the $j^{th}$ matcher classifies the result $k\in (c+1)$ for the match score $S_{j}$, it is denoted that for all instances the likelihood of $k$ being the correct class is $P_{k}$ and the likelihood of $k$ not being the actual class is (1-$P_{k}$). The induced score/decision output is computed by multiplying $P_{kj}$ with the respective match score $S_{j}$ for the $j^{th}$matcher. This score is then utilized as basic belief assignment or mass $m_{j}(k)$ as defined in Eq. (16):
\begin{equation}
m_{j}\left ( k \right )= P_{kj}\cdot S_{j}
\end{equation}
where $j$= 1,2 corresponds to the two matchers; one for the output achieved through integrating two different matchers for protected iris modality and the other for the output obtained by integrating both of the matchers over protected fingerprint templates. In a similar way, $ m_{j}\left ( \neg k  \right )$; with $m\left ( \theta \right )$ = 1 indicates the measure of disbelief. After the evaluation of induced score, the mass of each evidence or classifier is combined iteratively as described in Eq. (17):
\begin{equation}
m_{final}=m_{1}\oplus m_{2}\oplus m_{3}\oplus m_{4}
\end{equation}
where $\oplus$ represents the Dempster rule of combination. Here, we need not have to deal with the computation cost associated with DS theory \cite{shafer} since verification involves only two classes (accept, reject).

\subsection{Verification}
The utmost decision output is attained by employing a threshold ($t$) to $m_{final}$ as defined in Eq.(18). In this way, a user's identity can be verified to be a genuine or an imposter.  
\begin{equation}
Result=\left\{\begin{matrix}
Accept ;& \text{if} \ m_{final}> t \\ 
Reject ; & otherwise
\end{matrix}\right. \end{equation}

\section{Experimental results and analysis}
To perform successful multimodal verification, we present a number of experiments to demonstrate the performance of our proposed hybrid fusion framework encompassing score and decision level fusion. Subsection 5.1 describes the databases utilized in our work for experimentation. Subsection 5.2 narrates the experimental settings and performance metrics to quantify the results for each database. The performance of the proposed method is evaluated in Subsection 5.3. Subsection 5.4 presents baseline comparison to compare the performance of the method under the protected and unprotected scenario. Subsection 5.5 validates the achieved performance statistically. Next, we compare the proposed methodology with the other approaches in order to specifically measure the effectiveness and robustness of the proposed approach in Subsection 5.6. In section 6, we perform security analysis for our method and discuss the three major requirements needed for template protection as described in Section 1.

\subsection{Database}
We evaluate the performance of our method onto three virtual databases involving iris and fingerprint modalities. The virtual databases are created due to the underlying cost and efforts related to multimodal database acquisition. Virtual databases are constructed by pairing a user from one modality with a user from another modality. This pairing assumes that biometric traits of a user are independent. For iris, we use the CASIA V-3-Interval \cite{casia} database maintained by the Chinese Academy of Science and Multimedia university database (MMU1) \cite{mmu1}. The CASIA V-3-Interval database contains 2639 high-quality iris images from 249 users collected in two different sessions while MMU1 comprises of left and right iris images for 46 users. Considering left iris and right iris as different subjects, we find that there are 117 left and 121 right iris subjects from 348 total subjects from 249 users of CASIA V-3-Interval which contain at least 7 samples per subject. In MMU1 database, we consider a dataset of 92 users with 5 iris samples assuming left and right iris as a different subject. For the fingerprint modality, we use datasets DB1, DB2 of FVC2002 \cite{fvc} database containing a total of 800 images of 100 subjects with eight samples each.
\begin{itemize}
	\item First virtual database (Virtual\_A): Comprises of 100 subjects where iris images are randomly selected from 121 right iris subjects of CASIA V-3 Interval and fingerprint from FVC2002DB1 with 7 samples per subject.
	\item Second virtual database (Virtual\_B): Comprises of 92 subjects where iris images are selected from MultiMedia university version-1 (MMU1) database \cite{mmu1} and fingerprint from FVC2002DB2 \cite{fvc} with 5 samples per subject.
	\item Third virtual database (Virtual\_C): Comprises of 100 subjects chosen from 117 left iris subjects of CASIA V-3 Interval and FVC2002DB2 with 7 samples per user.
\end{itemize}

\subsection{Experimental settings}
After the generation of the protected template for iris and fingerprint, enrolled and query templates are compared to derive intra-class (i.e., genuine) scores and inter-class (i.e., imposter) scores. We adopt the FVC protocol to obtain the match scores. In FVC protocol \cite{fvc}, the first sample of a subject is compared with all remaining samples of the same subject to obtain genuine scores whereas the first sample of a subject is compared with the first sample of remaining subjects to measure imposter scores. First, the match scores from iris and fingerprint from different matchers are integrated by applying the proposed MC weighting method. As a result, the fused iris and fused fingerprint scores are obtained. Next, we evaluate induced score (decision output) from the computed match scores. Finally, we apply the DS theory of evidence to combined induced scores. The final decision output is compared with a predefined threshold to verify the user's identity. Further, the performance of our method is evaluated using the following metrics: 
\\
FMR: The probability of getting a positive comparison decision for an imposter \\
FNMR: The probability of getting a negative comparison decision for a genuine user \\
GMR: Can be measured as 1-FNMR \\
EER: The error rate where FMR and FNMR hold equality \\

\noindent Note that the focus of this work is the hybrid fusion (score and decision level) fusion for cancelable multimodal biometric, details of cancelable template generation is not reviewed here.

\subsection{Performance evaluation}
The larger context of this work is the hybrid fusion over the three described virtual multimodal databases for verification. The example images of the two modalities for Virtual\_A, Virtual\_B, and Virtual\_C databases are shown in Fig. 3.
\begin{figure}
	\centering  
	\subfigure{\includegraphics[width=0.20\linewidth]{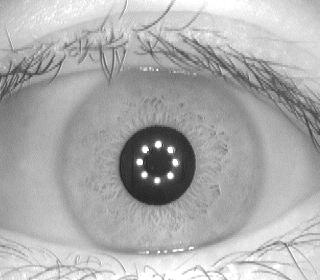}}
	\subfigure{\includegraphics[width=0.20\linewidth]{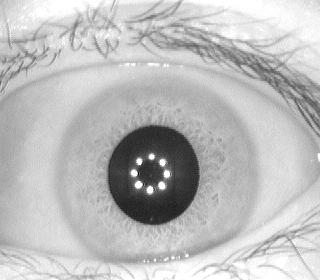}}
	\subfigure{\includegraphics[width=0.20\linewidth]{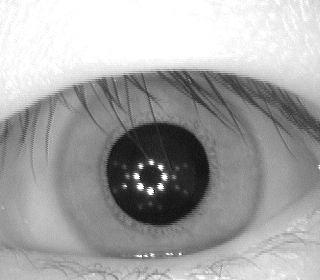}}
	\subfigure{\includegraphics[width=0.20\linewidth]{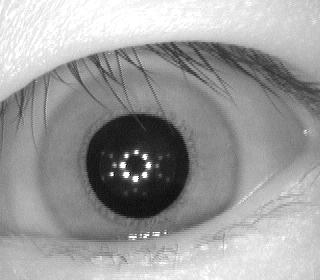}}
	
	\subfigure{\includegraphics[width=0.20\linewidth]{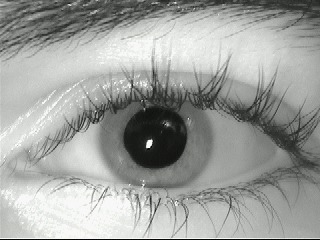}}
	\subfigure{\includegraphics[width=0.20\linewidth]{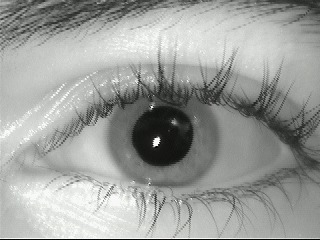}}
	\subfigure{\includegraphics[width=0.20\linewidth]{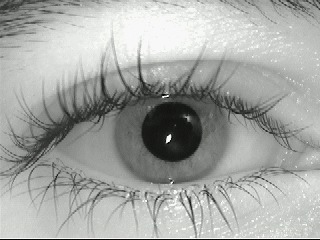}}
	\subfigure{\includegraphics[width=0.20\linewidth]{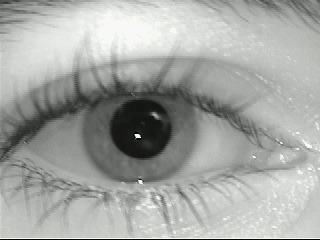}}
	
	\subfigure{\includegraphics[width=0.20\linewidth]{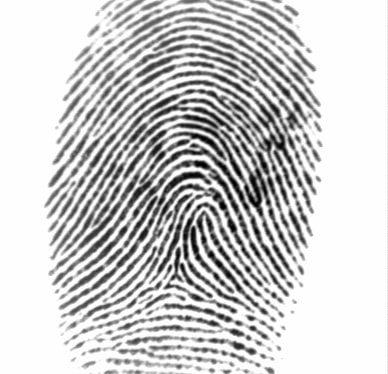}}
	\subfigure{\includegraphics[width=0.20\linewidth]{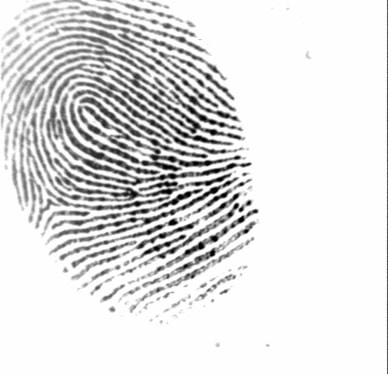}}
	\subfigure{\includegraphics[width=0.20\linewidth]{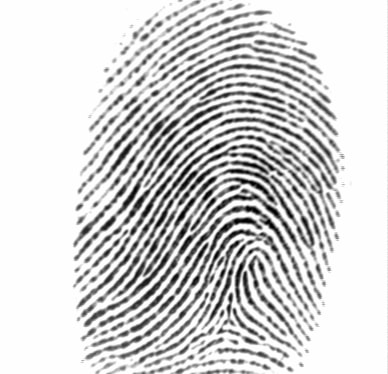}}
	\subfigure{\includegraphics[width=0.20\linewidth]{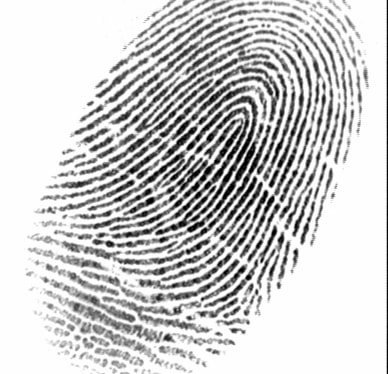}}
	
	\subfigure{\includegraphics[height=3cm,width=0.20\linewidth]{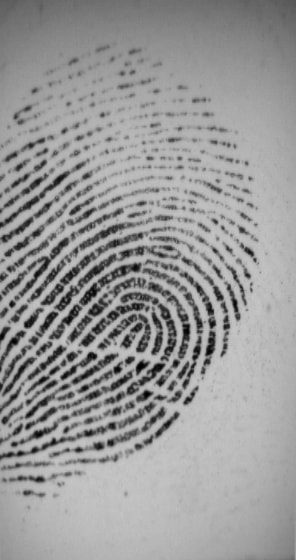}}
	\subfigure{\includegraphics[height=3cm,width=0.20\linewidth]{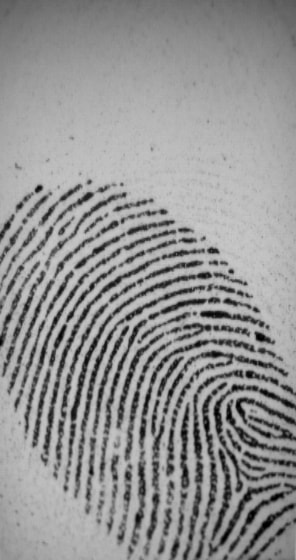}}
	\subfigure{\includegraphics[height=3cm,width=0.20\linewidth]{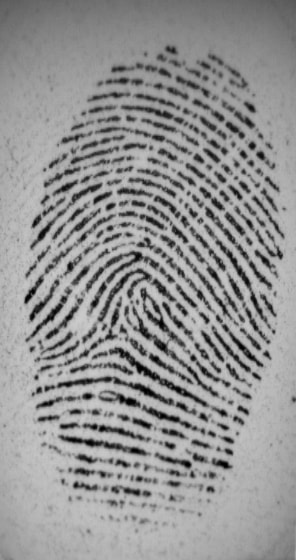}}
	\subfigure{\includegraphics[height=3cm,width=0.20\linewidth]{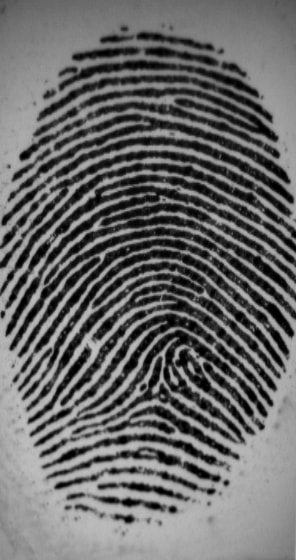}}
	\caption{First and second row indicate sample images from CASIA V-3.0 Interval and MMU1 database; Third and fourth row show the example images from FVC2002DB1 and FVC2002DB2, respectively.}
\end{figure}

After the generation of the protected templates for iris and fingerprint, stored protected and query protected template are compared with each other to calculate match scores. For iris modality, the match scores are derived using two matchers i.e., Hamming and Jaccard similarity whereas Dice and Cosine similarities are computed from fingerprint modality. To carry out the experimental evaluation, we perform training onto one set of each database. For each experiment, half of the total subjects are considered to train fusion retaining another half to test the performance of the proposed fusion framework. Further, we evaluate the match scores from the trained subjects. As a result, there are 1050 genuine scores and 1225 imposter scores for Virtual\_A and Virtual\_C multimodal chimerical database whereas we achieve 460 genuine scores and 1035 imposter score for Virtual\_B database respectively. The remaining half subjects are utilized for evaluation which results in the same number of genuine and imposter comparisons as mentioned before. Score level fusion is performed to fuse the match scores computed from different matchers corresponding to one modality. There is no parameter learning required for score fusion since we apply novel transformation based method (i.e., MCW) onto match scores. Next, we apply DS theory based fusion to combine integrated match scores corresponding to each modality. For each experiment, the training set is required first to find the parameters for decision fusion. In decision-level fusion, the parameters refer to the masses of the respective hypothesis. The masses have been computed for each induced score/decision output from different modality. These computed masses are combined using Eq. (6). Final verification decision is obtained by comparing the combined output with a pre-defined threshold (see Eq. 18). In DS theory based fusion, it is necessary to update the BBA if any new evidence or information is encountered. This updation is carried out using Eq. (8). Next, the evaluation is performed on test datasets for each virtual database. We evaluate EER values and Receiver Operating Characteristic (ROC) curves for unimodal and multimodal databases. Further, the performance in term of GMR @ 0.01\% FMR is computed since a biometric system deployed in a security application is considered to be efficient if it has low EER and high GMR at low FMR \cite{dec}.

The multimodal biometric performance of Virtual\_A is evaluated utilizing the scores obtained from CASIA V-3 Interval and FVC2002DB1. First, the performance for individual modalities (i.e., iris and fingerprint) taking part in fusion is evaluated. Next, we evaluate the performance of the multimodal biometric system. The ROC curve for the Virtual\_A multimodal database is shown in Fig. 4 which demonstrate the performance for the scores obtained in the unprotected and protected domain. From Fig. 4, it has been observed that the performance of the multibiometric system is better than that of a unimodal biometric system utilizing the proposed approach for both domains.

\begin{figure}[!htbp]
	\centering
	\includegraphics[width=\textwidth]{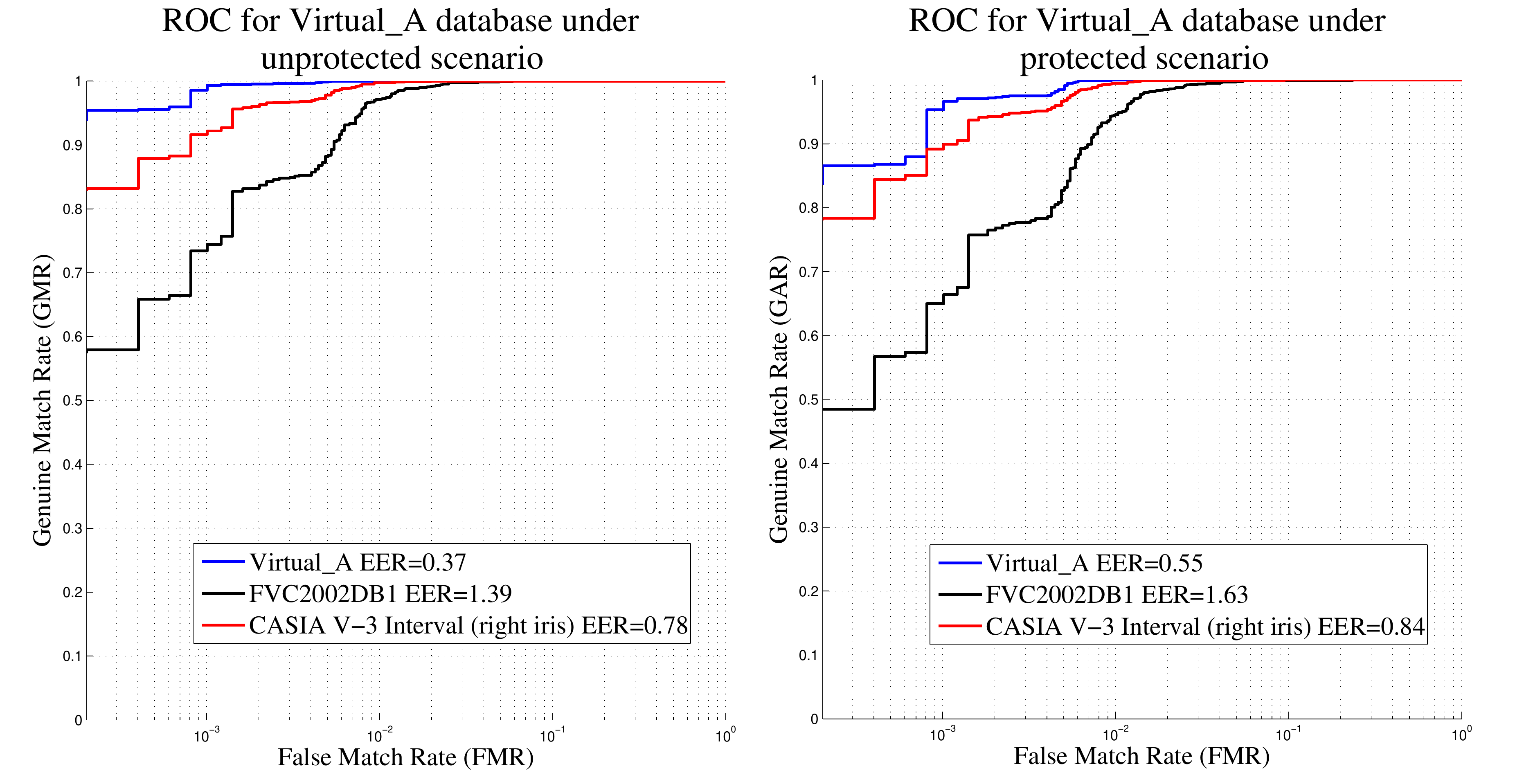}
	\caption{ROC curves for Virtual\_A database}
	\label{fig4}\end{figure}

In a similar manner, the Virtual\_B database comprising MMU1 iris and FVC2002DB1 is tested against our method. Figure 5 illustrates the ROC curve for the Virtual\_B database. We also demonstrate the ROC curves for individual modalities comprising Virtual\_B which clearly shows the superior performance for both of the scenarios.

\begin{figure}[!htbp]
	\centering
	\includegraphics[width=\textwidth]{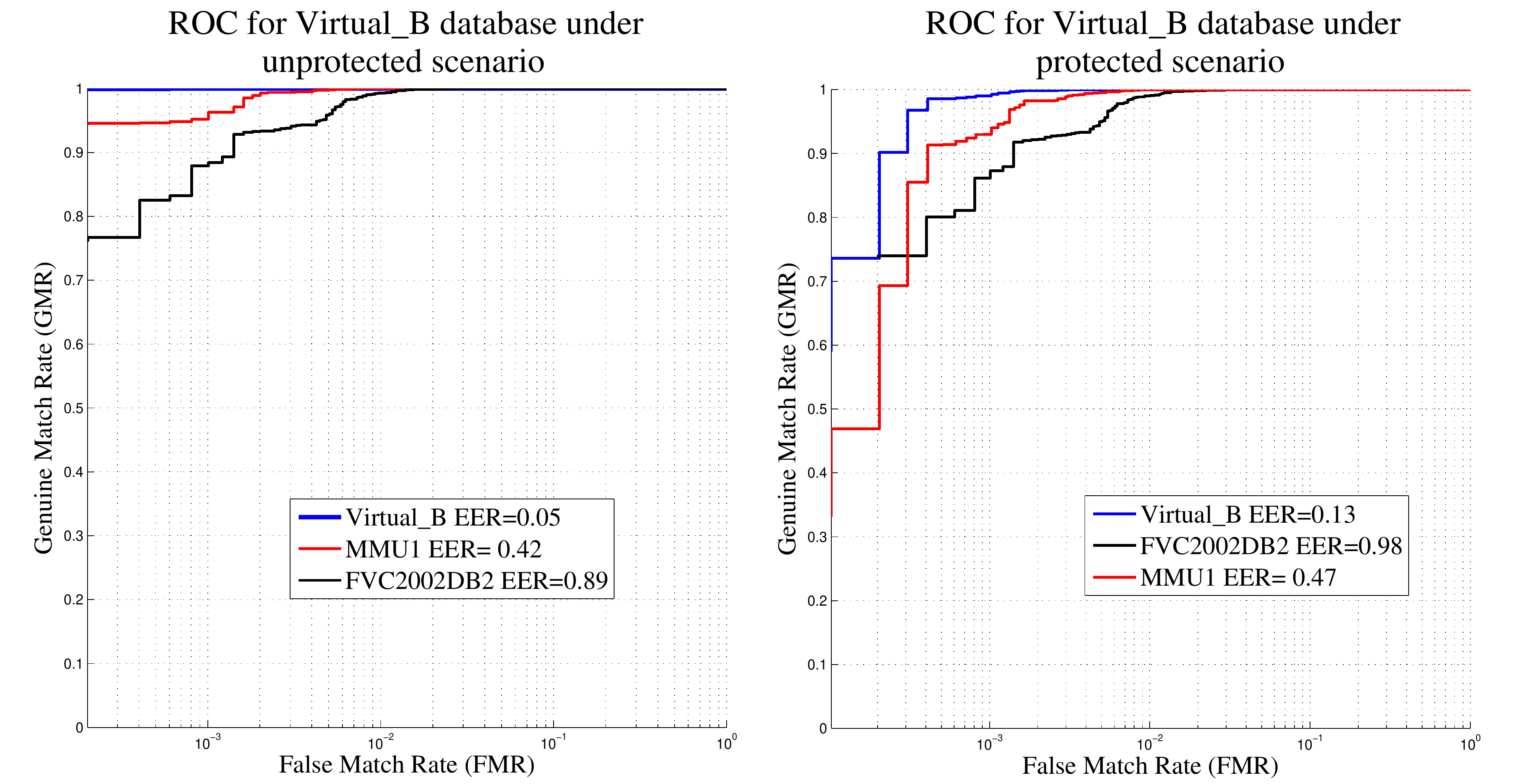}
	\caption{ROC curves for Virtual\_B database}
	\label{fig5}\end{figure}

Further, the performance for the Virtual\_C database comprising CASIA V-3 Interval iris and FVC2002DB2 is evaluated. The ROC curves for individual modality along with unprotected multimodal is shown in Fig. 6. It can be noticed from Fig. 6 that the proposed multibiometric system achieves better performance over the unimodal and unprotected multibiometric system for decisions obtained through original and cancelable biometric systems.

\begin{figure}[!htbp]
	\centering
	\includegraphics[width=\textwidth]{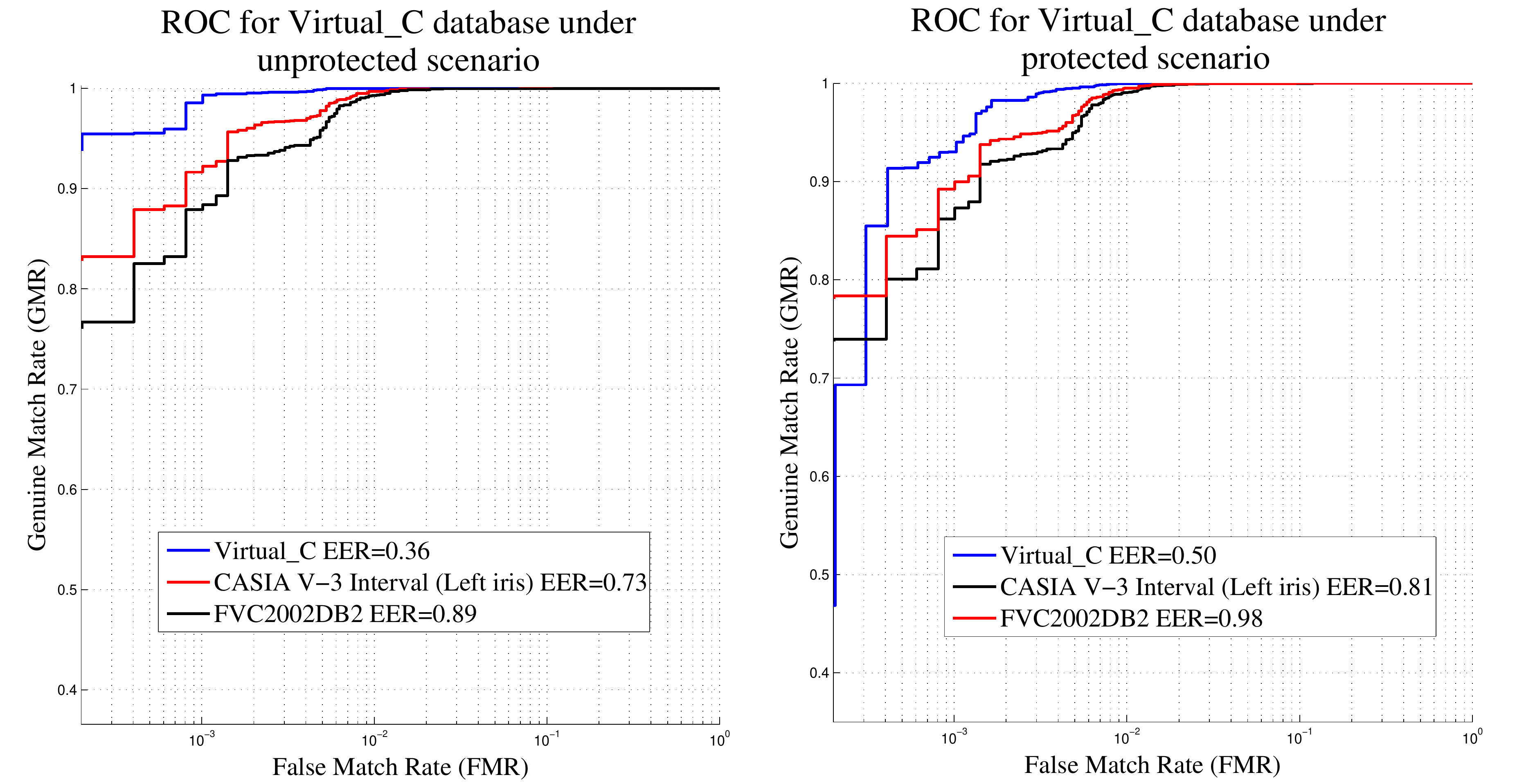}
	\caption{ROC curves for Virtual\_C database}
	\label{fig6}\end{figure}

In the proposed method, the performance is degraded by 0.32\%, 0.60\% and 0.28\% for Virtual\_A, Virtual\_B and Virtual\_C datasets, respectively under protected scenario as evident from the Fig. 4, Fig. 5 and Fig. 6. Therefore, we conclude that performance degradation produced by the cancelable transformation is very low. Further, we also evaluate the performance of our method in terms of GMR @ 0.01\% FMR and results are reported in Table 3 for Virtual\_A, Virtual\_B, and Virtual\_C databases, respectively. From Table 3, it is evident that the performance of the multibiometric system using the proposed method is better than that of other existing fusion schemes.
The performance for the Virtual\_B database is higher than that of Virtual\_A and Virtual\_C since there is a relative minimal overlap between the genuine and imposter score distributions. The extent of overlap is evaluated by decidability index $ d{}'$, which is defined as:
\begin{equation}
{d}'=\frac{|\mu_{1}-\mu_{2}|}{\sqrt{\frac{\sigma_{1}^{2}+\sigma_{2}^{2}}{2}}}
\vspace{0.3cm}
\end{equation}

where, $\mu_{1}$ and $\mu_{2}$ represent the genuine mean and imposter mean distributions, respectively; and the variances of the genuine and imposter score distributions are represented by $\sigma_{1}$ and $\sigma_{2}$ respectively. The value of $d{}'$ should be higher if the genuine and imposter distributions are more separable. We achieve the $d{}'$ of 2.74, 3.01 and 2.81 for Virtual\_A, Virtual\_B, and Virtual\_C databases respectively. The score distributions for all three chimerical databases are shown in Fig. 7. From Fig. 7, it is evident that the proposed fusion scheme achieves the optimal separation between genuine and imposter distribution for both the virtual databases.

\begin{figure*}[t!]
	$\begin{array}{rl}
	\includegraphics[height=5cm,width=0.5\textwidth]{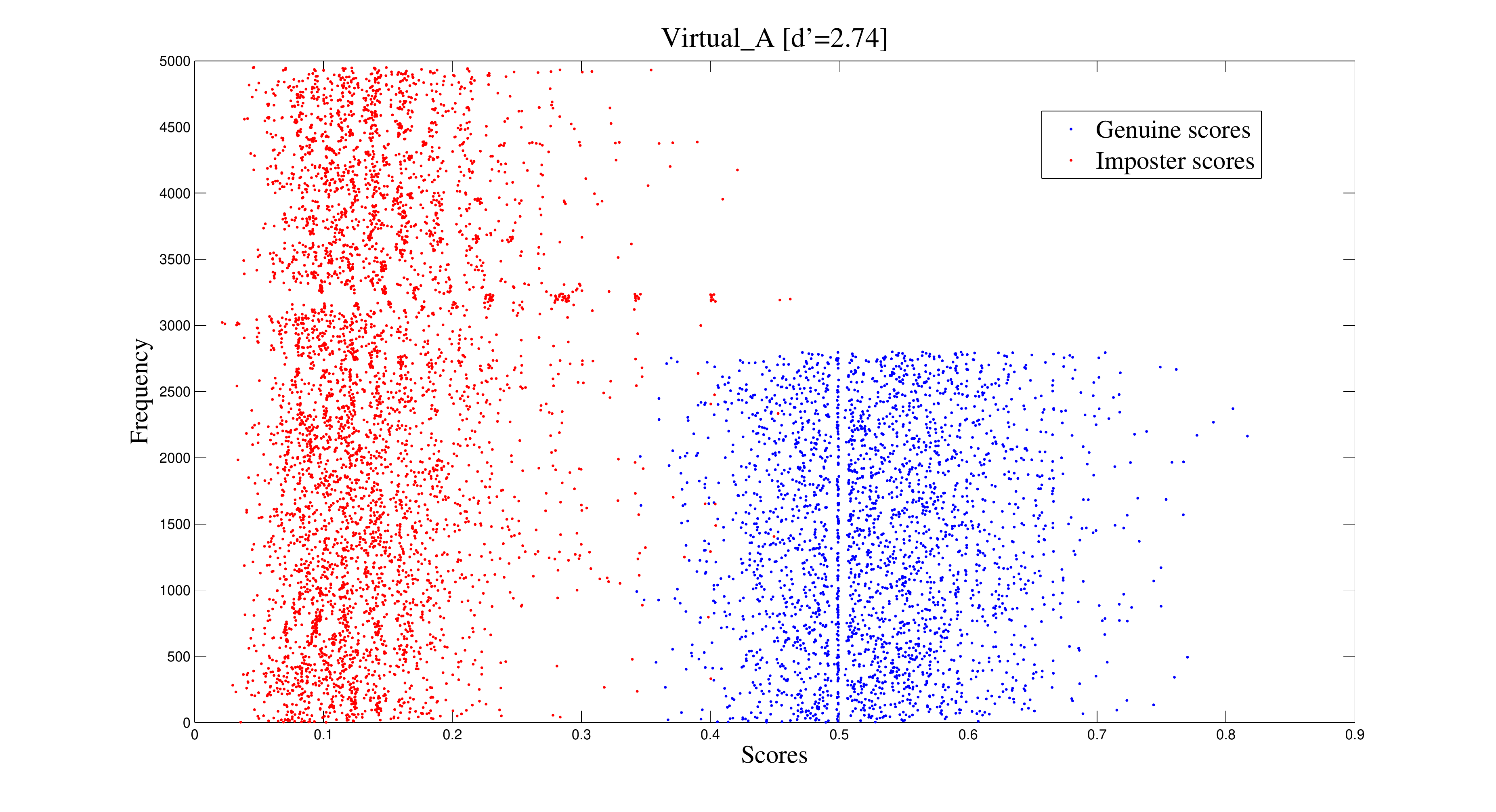} &
	\includegraphics[height=5cm,width=0.5\textwidth]{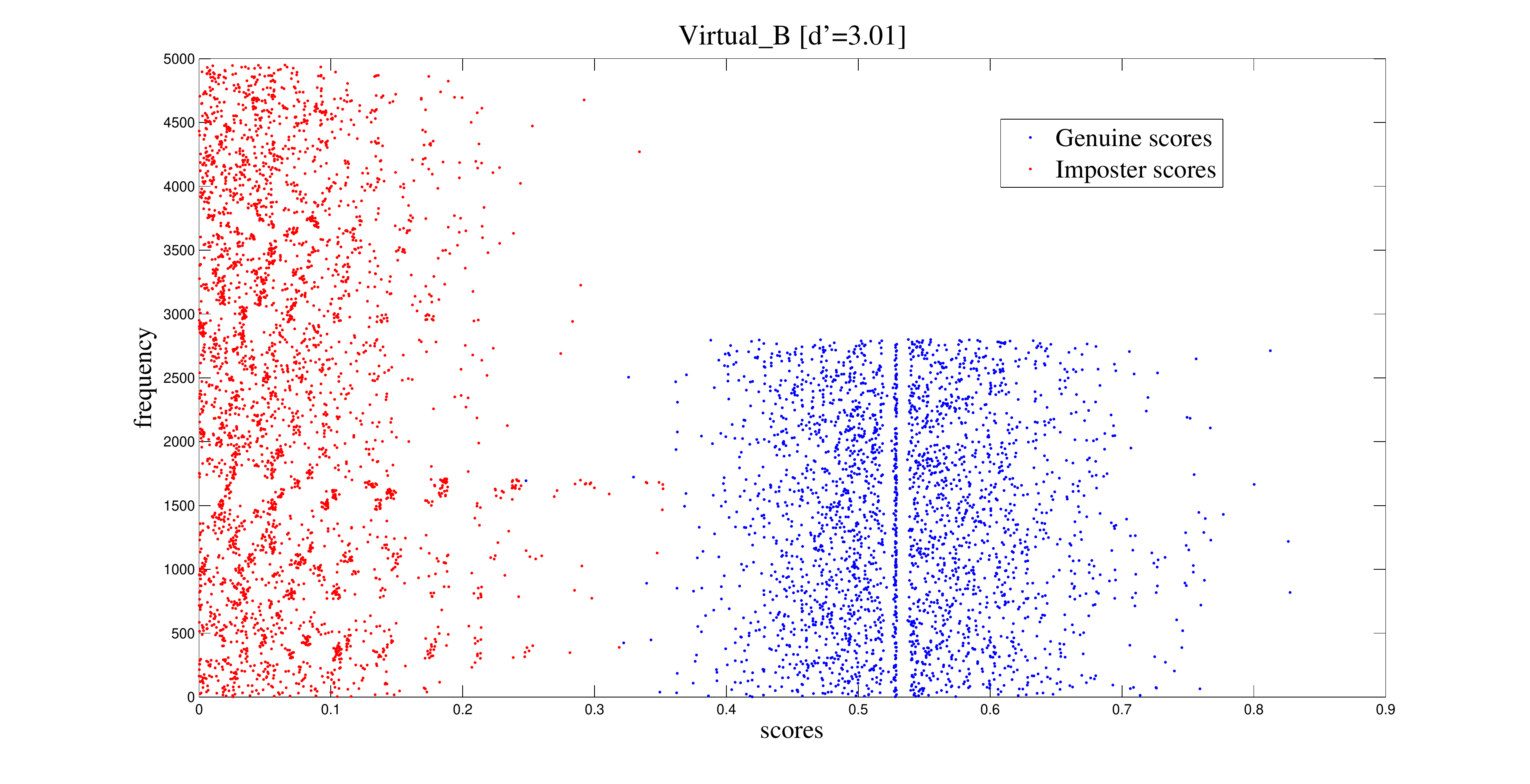}\\
	\multicolumn{2}{c}{\includegraphics[height=5cm,width=0.5\textwidth]{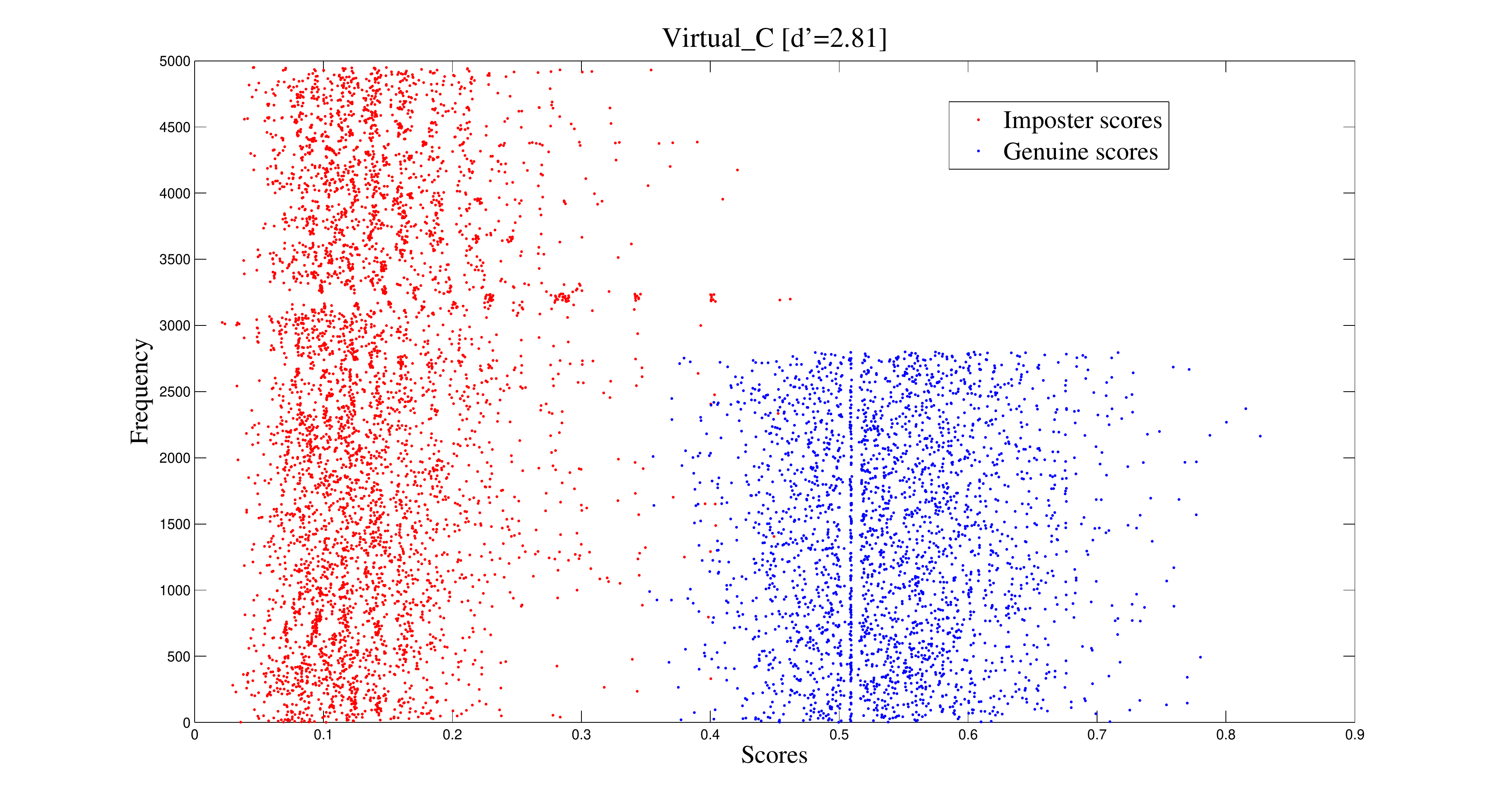}}
	\end{array}$
	\caption[Distribution curves of the fused matching scores]{\label{fig:label}Distribution curves of the fused matching scores}
\end{figure*}

\subsection{Baseline comparison}

Baseline comparison refers the comparison between the verification performance between protected multimodal and unprotected multimodal biometric system. In this work, we evaluate the performance obtained by combined scores from different modalities (fused iris and fused fingerprint) and final decision output under protected and unprotected scenarios. Further, we compare the performance of the proposed hybrid fusion framework with respect to the above-mentioned verification systems. Figure 8 represents the performance achieved in different scenarios for Virtual\_A, Virtual\_B, and Virtual\_C databases. Each of the figure comprise of (i) performances obtained by applying MCW over match scores from different matchers corresponding iris i.e. Score fusion [Protected iris], Score fusion [unprotected iris], Score fusion [Protected fingerprint], Score fusion [unprotected fingerprint], (ii) Hybrid fusion applied over iris and fingerprint modalities i.e. Hybrid fusion [Protected] and Hybrid fusion [Unprotected]. 

Figure 8 illustrates that hybrid fusion framework obtains 0.37 and 0.55 of EER under unprotected and protected scenario which is superior in comparison to fused iris unprotected (0.59), fused fingerprint unprotected (1.10), fused iris protected (0.67) and fused fingerprint unprotected (1.23) for the Virtual\_A databases. For the Virtual\_B database, the proposed method achieves an EER of 0.05 and 0.13 under unprotected and protected scenario which performs better than the results obtained in fused iris unprotected (0.23), fused iris unprotected (0.69), fused iris protected (0.31) and fused iris protected (0.77) verification systems. Similarly, we obtain superior results for Virtual\_C database also. Hence, it has been confirmed that hybrid fusion framework outperforms over individual score fusion systems for all three databases. 

\begin{figure*}[t!]
	$\begin{array}{rl}
	\includegraphics[height=5cm,width=0.6\textwidth]{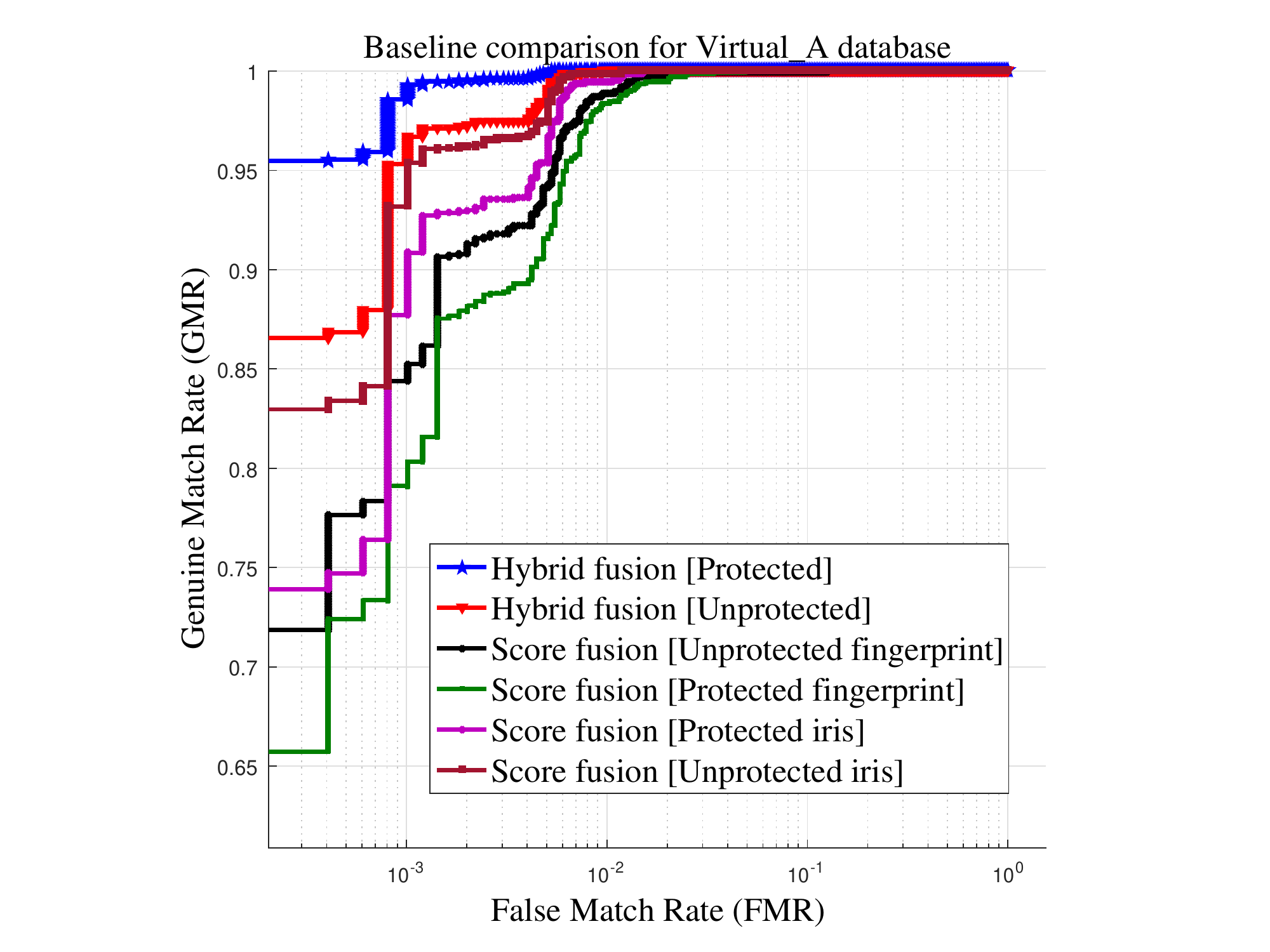} & \hspace{-1.5cm}
	\includegraphics[height=5cm,width=0.6\textwidth]{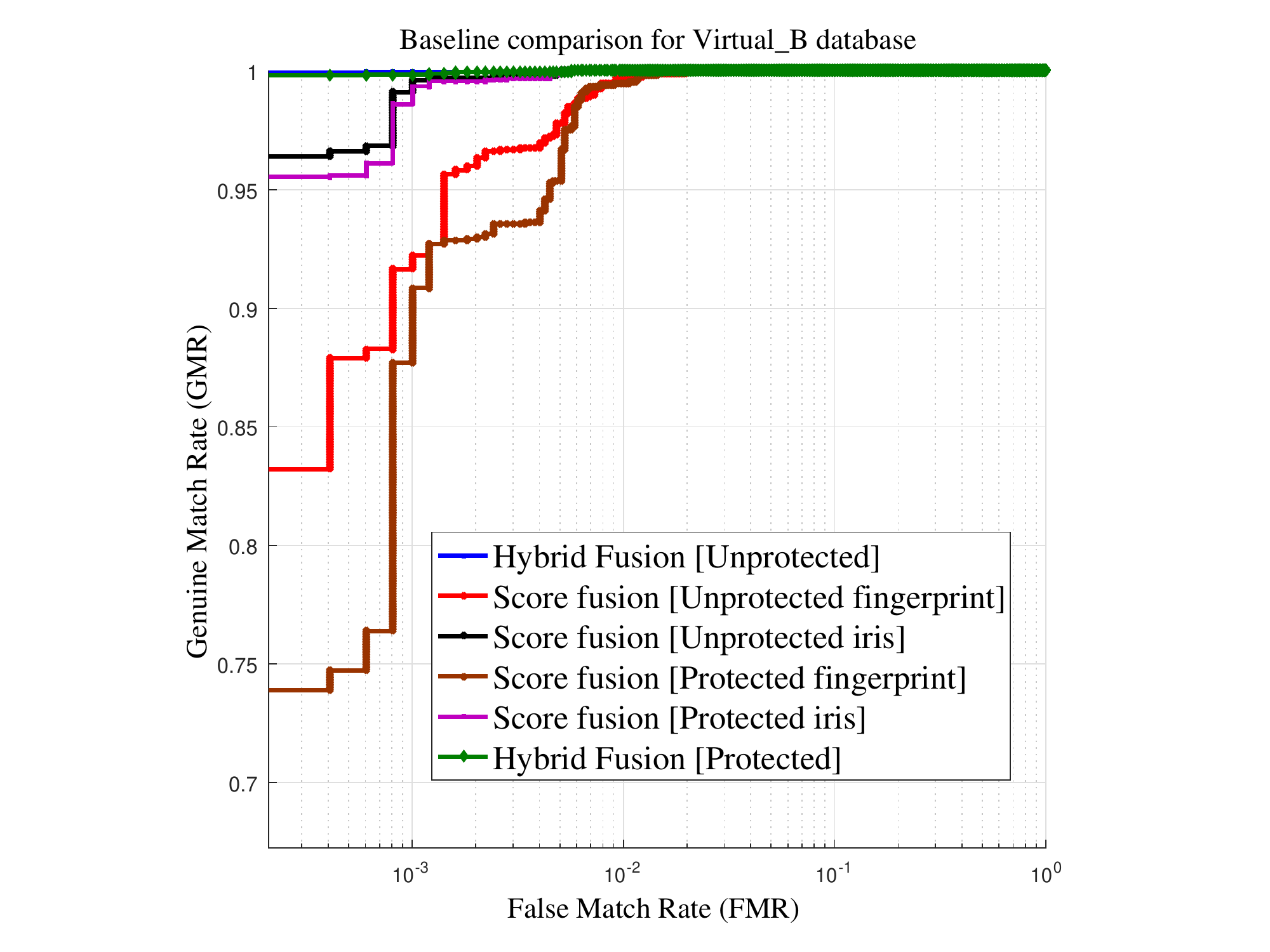}\\
	\multicolumn{2}{c}{\includegraphics[height=5cm,width=0.6\textwidth]{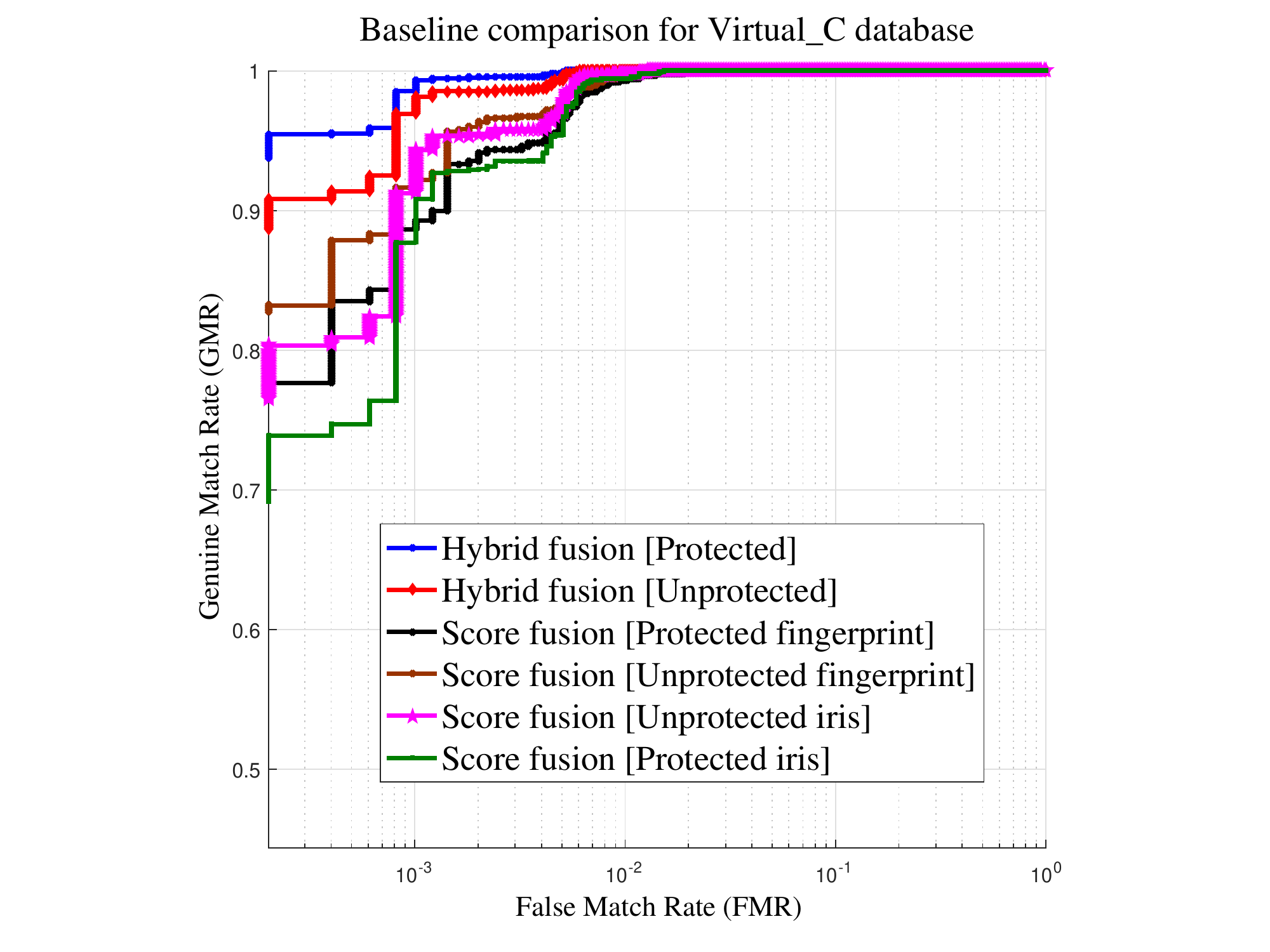}}
	\end{array}$
	\caption[Baseline comparison for the three databases]{\label{fig:label}Baseline comparison for the three databases}
\end{figure*}

\subsection{Statistical evaluation of proposed hybrid fusion method}
The performance of any biometric system is affected by the size of the database and the images comprising the database. ROC curves and verification performance are not enough to validate the overall performance of the multibiometric system. Hence, Bengio et al. \cite{bengio} presented a statistical test which utilizes half total error rate (HTER) and confidence interval (CI). Here, we test our method against these two parameters. HTER is computed as:
\begin{align*}
HTER=\frac{FMR+FNMR}{2}
\end{align*}

In order to compute CI around HTER, we look for the bound $ \sigma \times z_{\alpha/2}$. Here, $\sigma$ and $z_{\alpha/2}$ are defined as \cite{bengio}:
\begin{align*}
\sigma= \sqrt{\frac{FMR\left ( 1-FMR \right )}{4\cdot NI}+\frac{FNMR\left ( 1-FNMR \right )}{4 \cdot NG}}
\end{align*}
\begin{align*}
z_{\alpha/2}=\left\{\begin{matrix}
1.645 & \text{for} & 90\% \ CI \\ 
1.960 & \text{for} & 95\% \ CI \\ 
2.576 & \text{for} & 99\% \ CI
\end{matrix}\right.
\end{align*}

where, $NG$ and $NI$ represents the total number of intra-class comparisons and the total number of inter-class comparisons, respectively. We evaluate HTER and CI for both of the chimerical databases using the FMR and FNMR. The statistical evaluation is carried out at 0.01\% FMR and results are reported in Table 2. From Table 2, it has been observed that HTER lies between $0.02 \pm 0.05$ with 95\% confidence for the three chimerical databases which validates the achieved performance from our method.

\begin{table}[!ht]
	\centering
	\caption{Confidence interval around HTER of the proposed hybrid fusion method}
	\label{my-label}
	\begin{tabular}{|l|l|l|l|l|}
		\hline
		\multicolumn{1}{|c|}{\multirow{2}{*}{Database}} & \multicolumn{1}{c|}{\multirow{2}{*}{HTER (\%)}} & \multicolumn{3}{c|}{\begin{tabular}[c]{@{}c@{}}Confidence Interval (\%) \\ around HTER for\end{tabular}} \\ \cline{3-5} 
		\multicolumn{1}{|c|}{} & \multicolumn{1}{c|}{} & \multicolumn{1}{c|}{90\%} & \multicolumn{1}{c|}{95\%} & \multicolumn{1}{c|}{99\%} \\ \hline
		Virtual\_A & 0.54 & 0.02 & 0.033 & 0.043 \\ \hline
		Virtual\_B & 0.14 & 0.04 & 0.039 & 0.048 \\ \hline
		Virtual\_C & 0.52 & 0.03 & 0.042 & 0.050 \\ \hline
	\end{tabular}
\end{table}

\subsection{Comparison with other state-of-the-art fusion techniques}
To validate the performance of our method, we compare our proposed hybrid fusion scheme with other recent methodologies in literature. Besides hybrid fusion methods \cite{grover,hybrid2}, we include few other recent state-of-the-art fusion approaches based on score level \cite{ds1,acoajay,kabir,lmezai,scorelevel} and decision level fusion \cite{softbio,busch}. As described in performance evaluation, it can be observed that the proposed method performs optimally than the other approaches with respect to EER (see Figure 4-6). The superior performance is due to the extent of overlap ($d^{'}$) i.e. separability between the genuine and imposter distributions, as shown in Fig. 7. This also proves that the proposed method is less sensitive to the outliers since the separability between distributions is significantly higher than the existing methods.

First, the proposed hybrid fusion method is compared with other existing hybrid decision fusion schemes proposed in \cite{grover,hybrid2}. In case of Virtual\_A database, the technique proposed in \cite{grover} performs better than the proposed method, but it involves complex evaluation for global error optimization using PSO. The decision fusion methods involve AND rule and OR rule-based fusion proposed by Kelkboom et al. \cite{busch} and Bayesian classifier fusion proposed by Sadhya et al. \cite{softbio}. Table 3 reports the EER and GMR @ 0.01\% FMR, obtained using the proposed and existing weighting techniques. From the reported results in Table 3, it has been observed that the performance of AND rule and OR rule combination methods gets degraded in case the individual classifiers does not perform well. Hence, these two methods are rarely recommended in practice. Additionally, it can be analyzed from Table 3 that the proposed hybrid fusion outperforms the individual score level methods \cite{ds1,acoajay,kabir,lmezai,scorelevel}. The proposed hybrid multi-biometric system (i.e., cancelable iris - cancelable fingerprint system), provides lower EER and higher GMRs @ 0.01\% FMR than a majority of the existing techniques. Also, the best performance in terms of EER (i.e. 0.55, 0.13 and 0.50) and GMR @ 0.01\% FMR (i.e. 99.29\%, 99.70\% and 99.33\%) are achieved using the proposed method for the three virtual multimodal databases. Also, it is confirmed that the performance is enhanced by (48\%,66\%), (72\%,86\%) and (49\%,38\%) over unimodal cancelable systems for Virtual\_A (iris, fingerprint), Virtual\_B (iris, fingerprint), and Virtual\_C (iris, fingerprint) databases, respectively.

\begin{table}[!htbp]
	\centering
	\caption{Performance comparison of the proposed method with existing fusion approaches (in \%)}
	\label{compare}
	\resizebox{\textwidth}{!}{%
		\begin{tabular}{|c|c|c|c|c|c|c|}
			\hline
			\multirow{3}{*}{Methods} & \multicolumn{6}{c|}{Performance ( EER,GMR @0.01\%)} \\ \cline{2-7} 
			& \multicolumn{2}{c|}{Virtual\_A} & \multicolumn{2}{c|}{Virtual\_B} & \multicolumn{2}{c|}{Virtual\_C} \\ \cline{2-7} 
			& unprotected & protected & unprotected & protected & unprotected & protected \\ \hline
			\multicolumn{7}{|c|}{Score level fusion methods} \\ \hline
			Dwivedi et al. \cite{scorelevel} & 0.49, 99.59 & 0.69,98.89 & 0.09, 99.97 & 0.17, 99.64 & 0.45, 99.49 & 0.61, 99.25 \\ \hline
			Kabir et al. \cite{kabir} & 0.47, 99.44 & 0.62, 98.73 & 0.11, 99.80 & 0.17, 99.59 & 0.59, 99.12 & 0.71, 98.50 \\ \hline
			Nguyen et al. \cite{ds1} & 0.84, 98.81 & 1.12, 98.63 & 0.37, 99.49 & 0.45, 99.29 & 0.62, 99.28 & 0.79, 98.97 \\ \hline
			Kumar et al. \cite{acoajay} & 0.69, 99.18 & 0.78, 98.91 & 0.29, 99.58 & 0.41, 99.34 & 0.65, 99.30 & 0.83, 98.81 \\ \hline
			Mezai et al. \cite{lmezai} & 0.95, 98.85 & 1.19, 98.71 & 0.87, 98.99 & 1.10, 98.79 & 0.73, 99.18 & 0.89, 99.09 \\ \hline
			\multicolumn{7}{|c|}{Decision level fusion methods} \\ \hline
			Kelkboom et al. \cite{busch} & 1.52, 98.39 & 1.72, 98.13 & 0.97, 98.93 & 1.28, 98.70 & 0.78, 99.10 & 0.95, 98.91 \\ \hline
			Kelkboom et al. \cite{busch} & 1.41, 98.51 & 1.62, 98.32 & 0.81, 99.02 & 1.04, 98.83 & 0.69, 99.21 & 0.85, 99.01 \\ \hline
			Sadhya et al. \cite{softbio} & 1.01, 98.87 & 1.23, 98.70 & 0.55, 99.35 & 0.64, 99.23 & 0. 53, 99.38 & 0.67, 99.19 \\ \hline
			\multicolumn{7}{|c|}{Hybrid fusion methods} \\ \hline
			Grover et al. \cite{grover} & 0.34, 99.55 & 0.52, 99.39 & 0.09, 99.90 & 0.15, 99.81 & 0.42, 99.36 & 0.60, 99.27 \\ \hline
			Tao et al. \cite{hybrid2} & 0.52, 99.32 & 0.68, 99.07 & 0.19, 99.76 & 0.27, 99.61 & 0.51, 99.41 & 0.68, 99.24 \\ \hline
			\textbf{Proposed fusion} & \textbf{0.37, 99.64} & \textbf{0.55, 99.29} & \textbf{0.05, 99.98} & \textbf{0.13, 99.70} & \textbf{0.36, 99.55} & \textbf{0.50, 99.33} \\ \hline
		\end{tabular}%
	}
\end{table}

%

\section{Security analysis}
In this section, we present a general security model with all the components to perform exhaustive security analysis. For reader's clarity, we also describe each assumption taken into account for the entities associated with the verification procedure providing a more general perspective of how the multimodal fusion framework deals with different threats or privacy invasion attempts. An explanatory diagram (see Fig. 9 (left)) illustrates the verification procedure adopted for an unprotected scenario for two entities:

\noindent \textit{Client:} The client performs data acquisition, feature extraction and represents the features in the form of verifiable templates. Next, it computes the similarity score between the query and stored template. Finally, user's identity is verified based on a predefined threshold.

\noindent \textit{Server:} The server maintains the true biometric template for each user present in the database and shares these templates with the client for verification. To strengthen the privacy of a user, the server must send client's biometric data without pulling any other information and protect the biometric information stored in the database simultaneously.

\begin{figure}[!htbp]
	\centering
	\includegraphics[height=6cm,width=\textwidth]{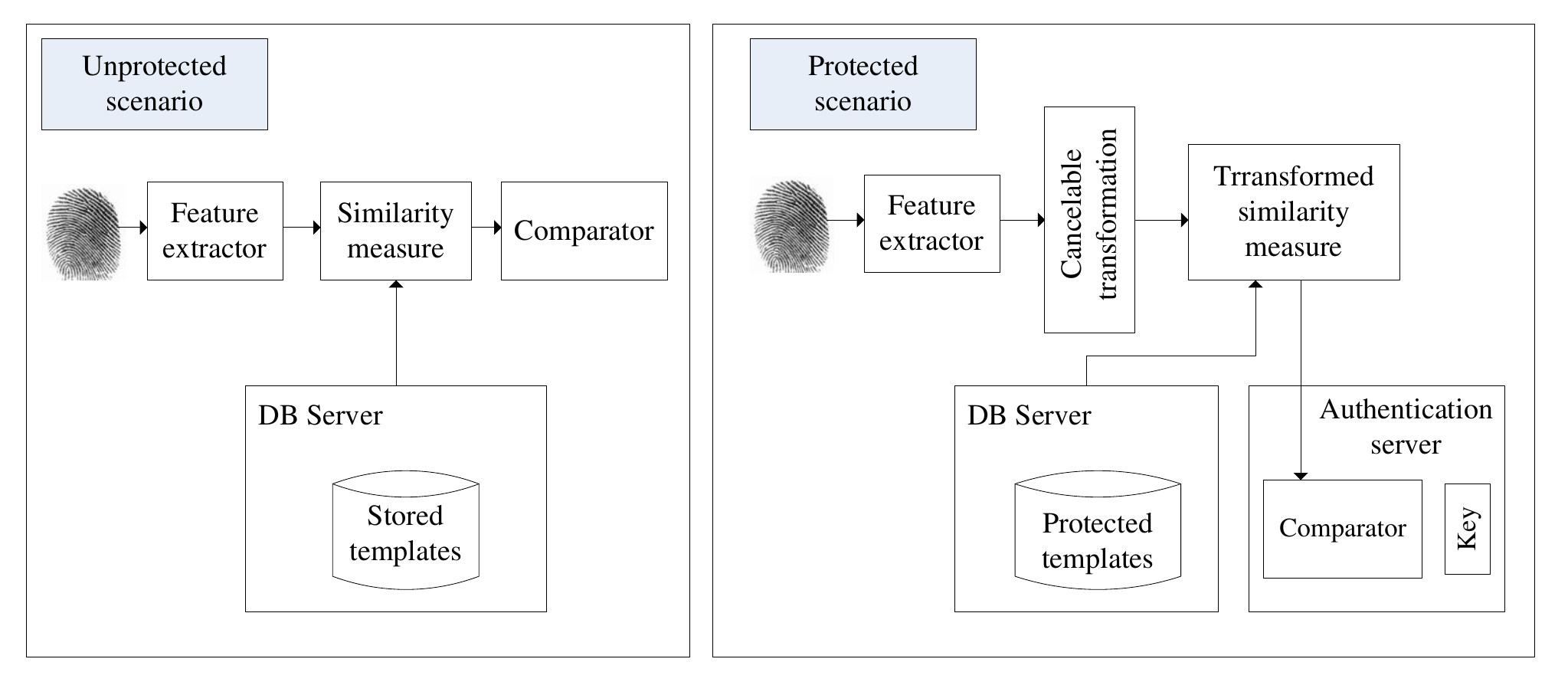}
	\caption{Unprotected vs. protected biometric verification}
	\label{fig2}\end{figure}

In contrast, a different security model is utilized for verifying protected biometric template is shown in Fig. 9 (right). In the protected scenario, all the biometric information which is either stored or communicated between client and server are transformed (i.e., protected). Hence, the mentioned entities play the following roles:

\noindent \textit{Client}: The client first acquires the data and extracts the features. Next, it applies a cancelable transformation to derive protected biometric templates and stores it onto DB server. 

\noindent \textit{DB server:} It contains the database consisting of only protected templates and shares these templates with the client for verification.

\noindent \textit{Authentication server:} It comprises the user-specific key and comparator. Also, it computes the final verification decision by comparing stored and query template.

The following assumptions are taken into account to perform secure authentication in a multi-biometric framework:
\begin{itemize}
	\item An imposter may get access to any one of the server but the DB server and authentication server would not intrigue.
	\item The client does not know the user-specific key hence it can neither extract the original template from the protected one nor the similarity score obtained through protected modalities assuming that the client serves honestly. As a result, there is no invasion possible of biometric information in the communication link. 
	\item Similarly, the authentication server would not be allowed to access to either the original or stored protected template avoiding any trace or instigate biometric data. Also, it is assumed that all involved entities adopt the protocol and thus the score evaluated by the clients are correct. 
\end{itemize}

\begin{figure}[!htbp]
	\centering
	\includegraphics[height=7.5cm,width=\textwidth]{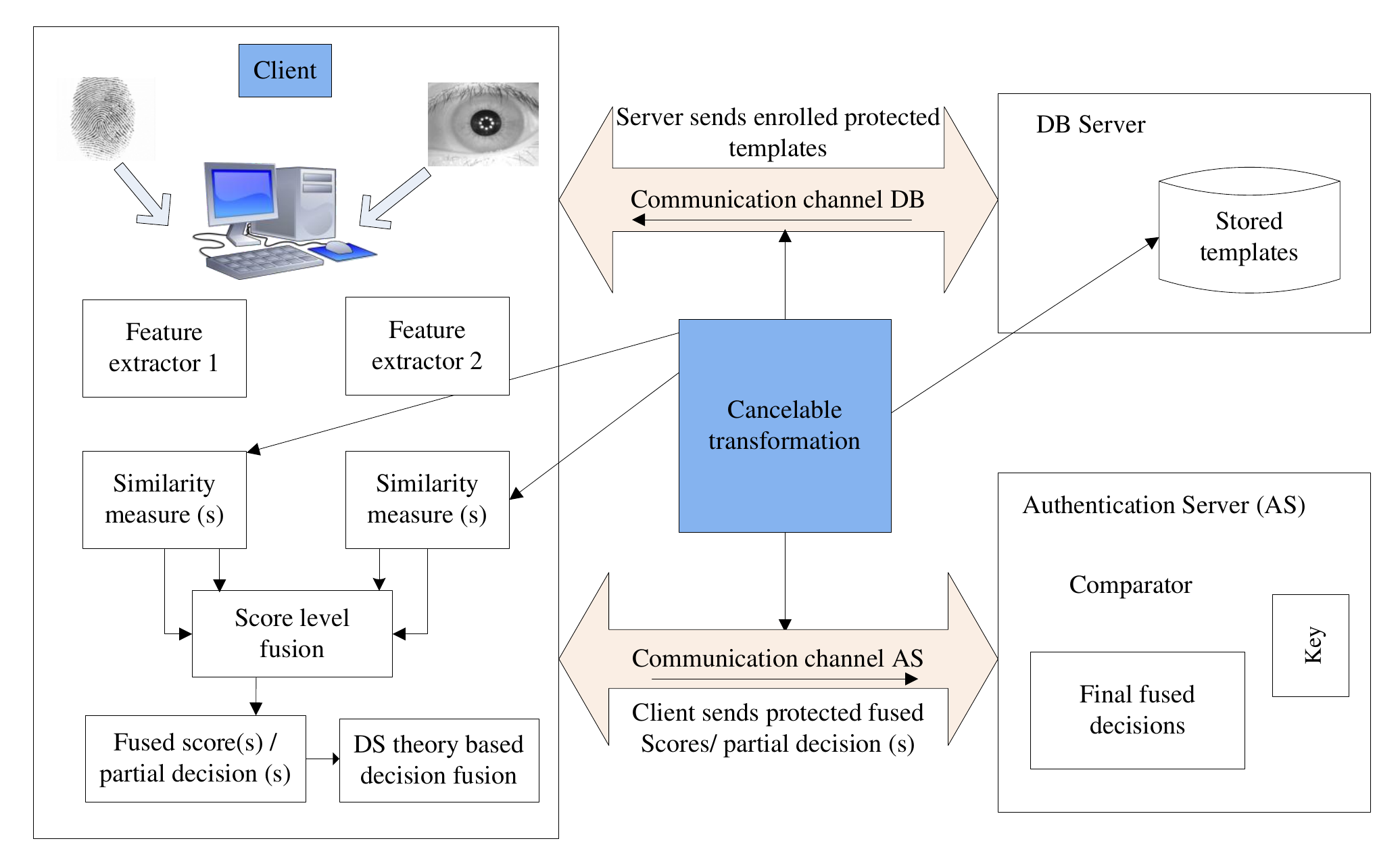}
	\caption{Security model: Hybrid fusion}
	\label{fig2}\end{figure}

Based on the security model illustrated in Fig. 10, the secure and privacy-preserving authentication in a multi-biometric fusion framework should exhibit the following requirements:
\begin{enumerate}
	\item The client alone should have access to the original biometric template 
	\item Only the protected template should be stored in the DB server and cannot be visible to any other entity
	\item The match score/ decision output cannot be transmitted as it may be utilized to launch inversion/ hill climbing attacks.
\end{enumerate}

To ensure the privacy protection, the authentication system should fulfill the three requirements, i.e. non-invertibility, diversity and revocability as described in Section 1.1. We will investigate these three criteria in the following subsections.

\subsection{Non-invertibility}
In our multimodal biometric fusion framework, only the protected template is shared/ communicated between DB server and client to compute the match scores/ decision outputs. Moreover, only the user-specific key is known to the authentication server. The authentication server can never get access to the stored and query protected template. Further, the client is not allowed to send any information to the extracted original template. Hence, it would be impossible for the client or any of the servers to trace any information related to template information since decisional composite residuosity in an NP-hard problem. Therefore, we can conclude that our approach meets the requirement of non-invertibility based on the ISO/IEC 24745 standard \cite{secu1}.

\subsection{Diversity}
In our approach, either a look-up table or random projection matrix or both of these can be altered to derive the numerous protected template corresponding to an instance of any subject. This ensures the criteria of diversity.

\subsection{Revocability}
To ensure potent revocability, the user-specific key can be altered to derive a new protected template and stored in the DB server. This way, the whole database could be re-secured with retransformed templates. This would avoid the impersonation of different users. These transformed templates for same or different subjects would be uncorrelated from each other. No information could be retrieved from these uncorrelated templates since the scores/ decision outputs from different modalities are computed in the protected domain.

In the proposed scheme, only the server is allowed to access the protected scores/ decision outputs from different modalities. Hence, inversion attack and Hill-climbing attacks \cite{secu2} are impossible to launch for an attacker, since he/she would not get the desired feedback to reconstruct the original template.

\section{Conclusion}
In this paper, we have proposed a novel hybrid fusion scheme for protected multi-biometric template verification based on score and decision level combination. Fusion at decision level is performed using DS theory of evidence and MCW weighting is employed to combine scores from different matchers corresponding to each modality. MCW weighting does not involve any learning incurring minimal computation complexity and DS theory exhibit a signification performance improvement thereby avoiding the uncertainty present in the matchers making ie efficiently applicable in military and government's security applications. Fusing the output of different matchers at the score or decision level allows the freedom to choose and evaluate any feature extraction or matching algorithm. In our method, score normalization is not required at any stage since the utilized matchers provide the scores already in the range [0,1]. In theory, the experimental evaluation carried out over three virtual databases depicts that the proposed fusion method will always outperform over the unibiometric authentication, and in practice, it also attain performance improvement better than the existing hybrid fusion and other conventional fusion schemes for multibiometric verification. Also, the performance evaluation showed that verification could be carried out in the transformed domain with no degradation. Further, the security analysis of our work ensures that our approach fulfills the desired characteristics of non-invertibility and revocability for template protection schemes by preserving the recognition accuracy. It is hoped that the proposed approach would be tested onto large databases containing 1000 subjects with more than two modalities. Additionally, we are also focusing on sequential and parallel decision level fusion for protected and unprotected multimodal biometric systems in future.
\begin{acknowledgements}
The authors are thankful to SERB (ECR/2017/000027), Department of science \& Technology, Govt. of India for providing financial support. Also, We would like to acknowledge Indian Institute of Technology Indore for providing the laboratory support and research facilities to carry out this research. 
\end{acknowledgements}

\bibliography{references}   


\end{document}